%% file: main.tex
\documentclass[preprint,10pt,5p]{elsarticle}

\usepackage{dirtree}
\usepackage{tabularx}
\usepackage{color}
\usepackage{algorithm}
\usepackage[noend]{algpseudocode}
\usepackage{amssymb}

\definecolor{cyan}{rgb}{0, 0, 1}

\input{preamble}
\input{acronym}
\begin{document}

\begin{frontmatter}

%\title{CPU power management for soft-deadline periodic real-time %system: the limitations of Linux built-in strategies and an %RL-based DVFS approach}
%\title{RL-driven CPU Power Management for Deadline Constrained Applications with Explicit Temporal Encoding}
%\title{Temporal Encoding based RL-driven Power Management for Deadline Constrained Applications}
%\title{Deriving Power Management Governor for Deadline-Constrained Applications on Small Devices via Reinforcement Learning with Temporal Encoding}
\title{CPU Frequency Scheduling of Real-Time Applications on Embedded Devices with Temporal Encoding-based Deep Reinforcement Learning}
%\title{Studying Practical Kernel-Level DVFS Techniques for %Periodic Applications with Soft-deadlines}
%\title{Studying Practical Linux Kernel-Level DVFS Techniques for Deadline-Constrained Periodic Applications}
% Document begins\cortext here

\author{Ti Zhou}
\author{Man Lin\corref{cor1}}
\ead{mlin@stfx.ca}

\cortext[cor1]{Corresponding author.}

\address{Department of Computer Science, St. Francis Xavier University, Nova Scotia, Canada}

% \IEEEtitleabstractindextext{
\begin{abstract}

Small devices are frequently used in IoT and smart-city applications to perform periodic dedicated tasks with soft deadlines. This work focuses on developing methods to derive efficient power-management methods for periodic tasks on small devices. We first study the limitations of the existing Linux built-in methods used in small devices.
We illustrate three typical workload/system patterns that are challenging to manage with Linux's built-in solutions. 
We develop a reinforcement-learning-based technique with temporal encoding to derive an effective DVFS governor even with the presence of the three system patterns. The derived governor uses only one performance counter, the same as the built-in Linux mechanism, and does not require an explicit task model for the workload.
We implemented a prototype system on the Nvidia Jetson Nano Board and experimented with it with six applications, including two self-designed and four benchmark applications. Under different deadline constraints, our approach can quickly derive a DVFS governor that can adapt to performance requirements and outperform the built-in Linux approach in energy saving. \textcolor{black}{
On \textit{Mibench} workloads, with performance slack  
ranging from 0.04 s to 0.4 s, the proposed method can save 3\% - 11\% more energy compared to \textit{Ondemand}. AudioReg and FaceReg applications tested have  5\%- 14\% energy-saving improvement.
}
\textcolor{black}{We have open-sourced the implementation of our in-kernel quantized neural network engine. The codebase can be found at: \href{https://github.com/coladog/tinyagent}{https://github.com/coladog/tinyagent}.}

\end{abstract}

% % Note that keywords are not normally used for peerreview papers.
% }

\end{frontmatter}

% \begin{document}
% Create the title.
% \maketitle

% Example sections, name them
% according to specific needs.
% \input{sections/abstract}
% \input{sections/keywords}
\input{sections/introduction}

\input{sections/background}
\input{sections/profiling}
\input{sections/motivation}
\input{sections/method}
\input{sections/experimentation}
% \input{sections/results}
% \input{sections/conclusion}
% \input{sections/discussion}
% \input{sections/acknowledgement}

% Select the IEEEtran style
% \bibliographystyle{IEEEtran}
\bibliographystyle{elsarticle-num} 
% Include bibliography file
% \bibliography{IEEEabrv,references}
\bibliography{references}

\end{document}

%% file: preamble.tex
\usepackage[utf8]{inputenc}
\usepackage[T1]{fontenc}
\usepackage[swedish,english]{babel}

\usepackage{graphicx, float, subfigure, blindtext}

\newcommand\IEEEhyperrefsetup{
bookmarks=true,bookmarksnumbered=true,%
colorlinks=true,linkcolor={black},citecolor={black},urlcolor={black}%
}

\usepackage[\IEEEhyperrefsetup, pdftex]{hyperref}

\usepackage{mathtools}

\usepackage[nolist,nohyperlinks]{acronym}
\usepackage{cleveref}
\graphicspath{{images/}}

%% file: acronym.tex
\acrodef{IC}[IC]{Integrated Circuit}
\newacro{svm}[SVM]{Support Vector Machine}

%% file: sections/introduction.tex
\section{Introduction}
\subsection{The Context of Energy Saving Problem}
Soft-deadline periodic real-time systems are commonly seen in many IoT/CPS/smart city/wearable computing systems  to provide ubiquitous and rich services.
The following are some sample systems reported in IEEE IoT Magazine.

\begin{itemize}
    \item \textbf{Smart dairy farm}: deploying sensors on cows to collect the biological information for the purpose of classifying their status \cite{taneja2019connected}.
    \item \textbf{Pest detection in precision agriculture}: using cameras to photograph the crop to detect the location of pest \cite{brunelli2019energy}.
    \item \textbf{Covid-19 screening and detection}: putting sensors on drones to collect biological information and  detect Covid-19 infection \cite{chintanpalliiomt}.
    \item \textbf{Smart irrigation}: collecting weather, crop growth, and soil conditions to analyze and predict whether the soil needs irrigation \cite{togneri2019advancing}.
    \item \textbf{Distance violation detection}: mounting cameras on vehicles to  detect distance violation \cite{sahraoui2020deepdist}.
\end{itemize}

These systems normally consist of multiple small devices filled with sensors/network calls and pre-defined periodically running  workloads to be completed before the next \textcolor{black}{task period}. Besides 
the performance requirement, {\it low power consumption} is  another important QoS requirement for such small devices for the battery life. 

Our goal in this work is to derive an adaptive model-free method that can save energy for such types of applications (Soft-deadline periodic CPS applications)  running on small devices that have limited computing capacity.

\begin{figure}
	\centering
	\includegraphics[width=0.5\textwidth]{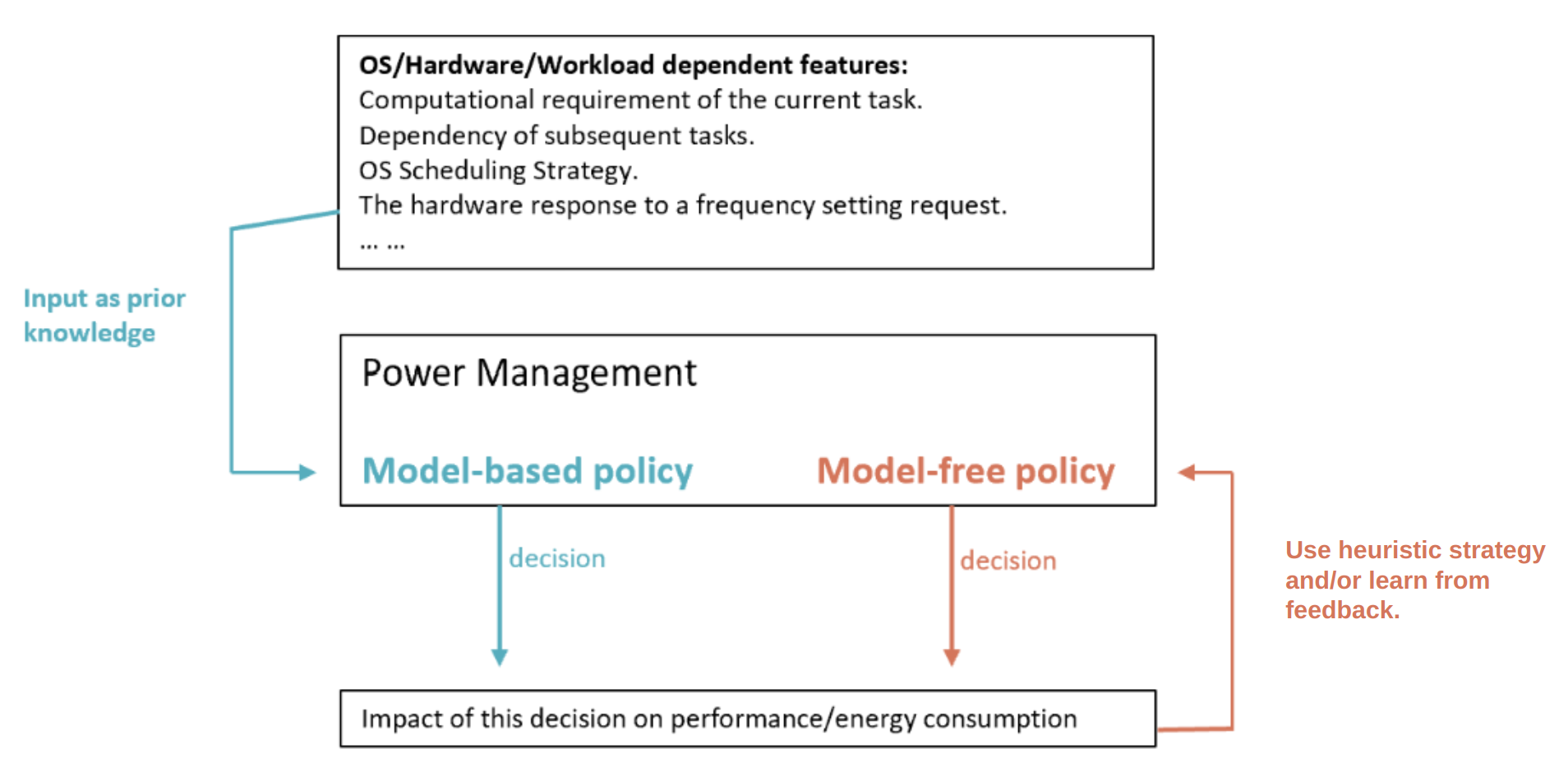}
	\caption{Model-based and Model-free Power Management policy.}
	\label{fig:exp2}
\end{figure}

\subsection{Model-based or Model-free Energy Saving Method?}
Modern small-device computing mainly relies on low-power CPUs. 
Dynamic Voltage and Frequency Scaling (DVFS) technology, which tunes the CPU's voltage (V) and frequency (f) on-demand, is a popular way to address such needs.
It is a classical problem in the real-time system community to schedule the CPU performance as energy-efficient as possible while satisfying the computational performance need. Many research works have been performed in this area.

Power management algorithms can be classified into two categories: model-based and model-free. Their differences are shown in Fig.~\ref{fig:exp2}. Model-based algorithms need a specification of the system with prior knowledge of the tasks \cite{reghenzani2021multi} \cite{ranjbar2020power} \cite{zhu2003scheduling} \cite{bhuiyan2018energy} \cite{bhuiyan2020energy} \cite{guo2019energy} \cite{saifullah2020cpu} \cite{paolillo2014power} \cite{huang2022energy} \cite{Li2015energy} \cite{Qiu2012threephase} \cite{sobhani2021realism}, including the attributes of the tasks such as worst-case execution time (WCET) \cite{ranjbar2020power} \cite{bhuiyan2018energy} \cite{sobhani2021realism}, deadline \cite{ranjbar2020power} \cite{bhuiyan2018energy} \cite{bhuiyan2020energy} \cite{guo2019energy} \cite{saifullah2020cpu} \cite{paolillo2014power} \cite{huang2022energy} \cite{Li2015energy} \cite{Qiu2012threephase}, \textcolor{black}{task period}\cite{zhu2003scheduling} \cite{bhuiyan2018energy} \cite{bhuiyan2020energy} \cite{paolillo2014power} \cite{Li2015energy}, possible priorities \cite{zhu2003scheduling} \cite{bhuiyan2020energy}, or the relations of the tasks specified as a DAG task graph with the communication cost and precedence order of the tasks \cite{reghenzani2021multi} \cite{ranjbar2020power} \cite{bhuiyan2018energy} \cite{bhuiyan2020energy} \cite{guo2019energy} \cite{saifullah2020cpu} \cite{huang2022energy} \cite{Li2015energy} \cite{Qiu2012threephase}. The mathematical model of the tasks, together with the machine architecture features, are used to find an optimal strategy for power management. On the other hand, a model-free DVFS algorithm does not require the input of task-specific information. They either %run algorithms
make decisions on frequency scaling based on some predefined assumptions (e.g. CPUFreq governors in Linux assume the future utilization is the same as the current CPU load measured by the performance counter), or collect feedback from the system during operation to adapt to the characteristics of the environment \cite{jung2010supervised} \cite{conradihoffmann2021online} \cite{park2021interpretable} \cite{das2015workload} \cite{tan2009adaptive} \cite{liu2010enhanced} \cite{wang2016model} \cite{ul2015hybrid} \cite{ramegowda2022energy} \cite{wang2021online} \cite{shafik2015learning} \cite{panda2022energy}.

The Model-based approach is limited to systems with a known explicit model. The detailed timing behavior of the tasks running on a specific target machine needs precise knowledge of the tasks and the architecture of the target machine, which could affect the timing behavior of the tasks. The system model (tasks + device) is normally hard to obtain. Therefore, such methods are only adopted for hard real-time systems that are safety-critical when a careful analysis of the system and tasks is necessary.

\subsection{Problem Statement and Proposed Approach}

Model-free DVFS methods are the common practice in general-purpose operating systems, such as Linux. Their power management policy can be applied to any task. We adopt a model-free power management method, given that many CPS applications are deployed to systems without prior knowledge of an explicit task model and device model.

We first look at the common structure (architecture) of a model-free DVFS algorithm, which includes the following.
\begin{enumerate}
    \item Use performance counters to construct the current system features.
    \item Use a model for inference.
    \item Select the next period's CPU frequency based on the model's output.
\end{enumerate}
The Linux built-in methods (\textit{Ondemand}, \textit{Conservative} and \textit{Schedutil}) are a classic application of the above architecture.
They sample the CPU utilization of the past period as a system feature, predict the constant computational demand for the next period, and then update the CPU frequency based on some heuristic rules.

Recent research efforts have focused on how to improve this architecture. Methods include: enabling more performance counters to build complex system features \cite{conradihoffmann2021online} \cite{park2021interpretable} \cite{wang2021online}, using powerful models for prediction \cite{conradihoffmann2021online} \cite{jung2010supervised} \cite{park2021interpretable} \cite{das2015workload} \cite{wang2016model}, designing better control rules \cite{conradihoffmann2021online} \cite{jung2010supervised} \cite{park2021interpretable} \cite{das2015workload}, or learning control policy based on reinforcement learning \cite{wang2016model} \cite{ul2015hybrid} \cite{ramegowda2022energy} \cite{wang2021online} \cite{zhou2021deadline}.

For general systems, where the arrival time of tasks is highly variable and diverse, improving power management strategies can be difficult. However, periodic systems are special from a temporal point of view, as they run a set of pre-defined tasks periodically. Can we exploit this feature to develop a better CPU power management policy for a system?

This work focuses on how to  {\bf deriving efficient model-free methods for periodic tasks with a soft-deadline running on small devices}.
\subsubsection{Study the Limitation of Existing Model-Free DVFS Governors through Profiling}
To avoid reinventing the wheel, our first step in approaching this problem is to study the behavior of existing Linux built-in governors through kernel-level profiling. We want to 
study if simply tuning the existing built-in governor will result in a better energy-efficient governor that is tailored for the periodic tasks with a soft deadline. In tuning the power management strategy for periodic soft real-time systems commonly found in contemporary embedded systems, we observe that the structure of existing DVFS governors can be ineffective for three frequently occurring system patterns. To be more specific,  CPU cores only have coarse-grained voltage/frequency level (limited) support, which is typical for small devices, cores can experience unbalanced load distributions, and tasks can have internal slack. 

This is mainly because existing built-in DVFS algorithms focus on the short-term computational characteristics of the system, whereas a good strategy for achieving an overall optimal solution often requires macroscopic knowledge: what has been computed in the past and what will be computed in the future.

This motivates us to find alternative methods to obtain a model-free governor rather than attempting to tune the existing governor for energy saving for the particular class of workload (soft-deadline periodic tasks) that we are interested in. The problem can be thought of as a sequence of decision-making of assigning a frequency to the CPU at each decision point. Reinforcement learning is a natural strategy to apply in the absence of an explicit model of the tasks to help with decision-making.

\subsubsection{Using Reinforcement Learning with Temporal Encoding to Derive Model-Free Governors}
A reinforcement learning approach will learn a DVFS inference model (a governor) through the feedback of the sequence frequency decisions, including its effect on the system load, timing, and energy consumption.  
Note that every frequency assignment in the sequence makes a difference in the amount of energy used at the end and how long it takes to complete the task. So, a pool of time series data will form the foundation of the learning process.

One of the most crucial design issues for reinforcement learning the state representation. 
Automatic encoding of features from raw input data has been the main focus of recent artificial intelligence.
In a previous work \cite{zhou2021deadline},  RNN was used to encode time series automatically. Until now, this approach has relied on complex computations and lengthy learning from large amounts of data. Another drawback of automatic encoding is its poor interpretability.

We choose an explicit encoding for the temporal information in this work to achieve higher interpretability and to ease the burden of model learning, which is important for small devices with limited computing resources.

\subsection{Contribution}
We choose Nvidia Jetson Nano 2GB board as our experiment testbed.

Our first contribution is to study the limitation of existing Linux built-in DVFS methods through profiling.
To analyze the CPU frequency control policy, we designed and implemented a low-overhead in-kernel profiler to collect the complete ms-level runtime data of the Linux CPUFreq governor.
With this profiler, we identified three scenarios when the Linux built-in DVFS methods are ineffective.
\begin{itemize}
  \item Target devices only support coarse-grained voltage (or frequency) levels, which are common for small devices. For example, both Raspberry Pi 4B+ and Nvidia Jetson Nano Board 2GB only support two V/f levels.

  \item Multi-core architectures have unbalanced CPU load distribution.
  
  \item There exist internal slacks within the workload caused by IO calls (sensors, cameras, microphones, network interactions, etc.).
\end{itemize}

Our second contribution is to design a reinforcement learning-based frequency governor under the CPUFreq framework with a temporal encoded system state to better address the above challenges.

\begin{figure}
	\centering
	\includegraphics[width=0.5\textwidth]{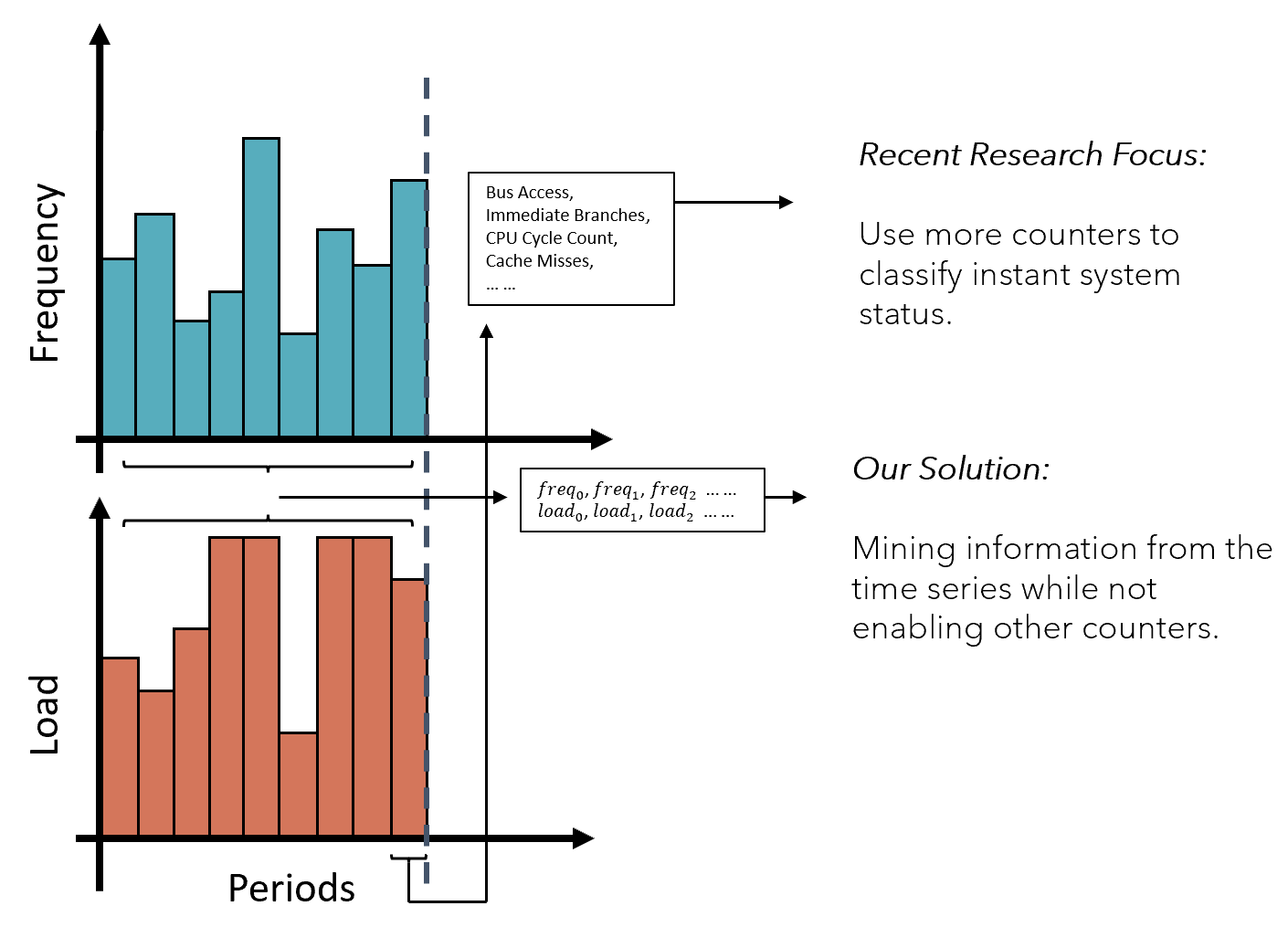}
	\caption{How to understand the computing demand for the next period?}
	\label{fig:exp1}
\end{figure}

We only use CPU load (utilization) for decision-making as standard built-in Linux governors. However, the instance information of CPU load (utilization) cannot provide sufficient distinction for the CPU to make different frequency scaling choices that consider the computation requirement of the current tasks. Therefore, the state construction of our method is based on the temporal sequence of the load instead of the load value at the previous instance, as shown in Fig.~\ref{fig:exp1}. The temporal sequence is encoded as a vector reflecting the progress of task execution for the RL governors to make a decision. The temporal encoding enables a reinforcement learning method to efficiently understand the workload from a macro perspective by mining the timing sequence even without the explicit model of the workload. The domain-assisted encoding is different from Standard RNN, where the encoding vector is learned, making on-device learning infeasible if the application workload sequence is long. Experiment results show that the encoding, together with the reinforcement learning method, is effective for finding good DVFS scaling strategies through on-device reinforcement learning.

The following summarizes the advantage of our energy-saving framework with on-device learning for periodic CPS applications with soft deadlines.
\begin{itemize}
 \item \textbf{High Interpretability}. We carefully designed the system state used in reinforcement learning to include features that intuitively contain valuable information.
 \item \textbf{Low Deployment Complexity}. Excessive training time due to large amounts of trial and error can lead to increased deployment complexity. We achieve low training time by properly designing the state to contain explicit and useful information from a human expert perspective to help the model quickly link the cause and the result. In our experiments, the proposed method learns a good DVFS policy with only three hundred workload runs.
 \item \textbf{Low Resource Overhead}. Similar to \cite{zhou2021deadline}, our work only implements the decision inference component at the kernel level. The learning component is implemented at the user level with data collected by the in-kernel profiler. Thus, the kernel state is only burdened with little inference overhead. 
 Our work further reduces the overhead by applying quantization \cite{jacob2018quantization}, with which the kernel can avoid floating-point calculation. 
 In our experiments, an inference of the proposed method takes only 
25.62 us with 1.479 GHz on average.

\end{itemize}

%% file: sections/background.tex
\section{Background}

% \subsection{CPU Power Model}
\subsection{Dynamic Power Consumption}

The dynamic power consumption $P_d$ of a CMOS circuit is determined by \cite{kaxiras2008computer}:

\begin{equation}
P_d = \alpha \times C \times V_{dd}^2 \times f,
\label{pd}
\end{equation}

where $V_{dd}$ is the supply voltage, $f$ is the clock frequency, $\alpha$ is the switching activity level, and $C$ is the capacitance of the circuit.

Supply voltage $V_{dd}$ and clock frequency $f$ are related as follows:

\begin{equation}
f \propto \frac{\beta(V_{dd} - V_{th})^2}{V_{dd}}, \label{Vddf}
\end{equation}

where $V_{th}$ is the threshold voltage, and $\beta$ is a technology-dependent constant.
For $V_{dd} \gg V_{th}$ and $\beta$ closed to 1, clock frequency $f$ is roughly proportional to $V_{dd}$.
In this case, the dynamic power consumption is proportional to $V_{dd}$ and $f$ through a cubic relationship:

\begin{equation}
P_d \propto V_{dd}^3 \propto f^3 \label{cubic}
\end{equation}

Dynamic power consumption reduces with the frequency following a cubic relationship, whereas execution time increases following a nearly linear relationship.
This property determines that, for the same task, it can be executed with less energy at a lower frequency/voltage.
Suppose only the frequency is reduced, but the voltage stays the same. Due to the reduced current in this scenario, the instantaneous power consumption is lower. However, since the running task will take longer to complete, the total amount of energy used to complete one task will not be lowered.

It is worth mentioning that even if the energy consumption cannot be reduced, the heat generation of the system will be reduced due to the decrease in the instantaneous power.
However, if hardware costs permit, it is desirable to regulate the voltage and frequency together.

\subsection{Static Power Consumption}

Static power consumption $P_s$ represents 20-40\% of the power budget of microprocessors in modern fabrication technologies \cite{kaxiras2008computer}, it is determined by:

\begin{equation}
P_s = I_{static} \times V_{dd}  \label{ps}
\end{equation}

$I_{static}$ is primarily due to subthreshold leakage current, and gate leakage current \cite{kaxiras2008computer}, which are affected by the supply voltage $V_{dd}$.
Lowering $V_{dd}$ can save both dynamic power consumption and static power consumption.
When the CPU is idle, lowering the CPU voltage can effectively save energy.
On Jetson Nano Board 2GB, when the system is idle, by setting the CPU to the lowest voltage, the board-wide power consumption (measured by a power meter) can be reduced by 36\%.

CPU Idle Time Management, which shuts down part of the CPU hardware function when idle, is another efficient way to reduce static energy consumption.
However, the more CPU functions are turned off, the more time and energy are required to switch back to a normal state.
Software-level algorithms need to be implemented to predict the idle duration of the CPU to select the appropriate idle state to enter.
Poorly designed idle control algorithms can waste energy and lose performance at the same time.

%% file: sections/profiling.tex
\section{Low Overhead Kernel Profiling}

The process of CPU frequency scaling can be viewed as an agent (frequency governor) observing the environment (the computing device managed by the OS that runs the work-
load) and taking actions accordingly (frequency scaling). In order to better comprehend the advantages and disadvantages of different policies, we want to depict the decision-making process, which can also enhance the interpretability of a learning-based solution.

Reading kernel data via default Linux support (for example, character file systems like \textit{sysfs}) or advanced tools (for example, \textit{perf} \cite{perf}) often involves reading a string from a buffer/file and then extracting data from it. The two built-in Linux CPU frequency governors (\textit{Ondemand} and \textit{Conservative}) typically perform 100 inferences per second by default \cite{cpufreqdoc}. 
Performing Perf-like \cite{perf} operations at such a high rate will put a non-negligible burden on the real workload of the system, which would cause the resulting profiling data to be meaningless.

% {\color{red}
Our objective is to profile the complete in-kernel CPU tuning data at the micro-second level while ensuring low latency.
Our solution involves inserting a profiler into the CPU governor that, at runtime, sends data directly to the kernel's data structures and writes data to the shared file system only at the end of the system run.

Each time the CPU governor makes an inference, our profiler collects necessary data into an array in DRAM.
Specifically, our profiler writes 42 bytes of data per inference. If the governor performs 100 inferences a second, then a 14.17MB array is sufficient to store all the information generated in one hour.
At the end of the system run, the profiler writes the collected data to the shared file system accessible to the user-state. 
The final write-out overhead barely has any effect on the runtime workload that is being profiled.

\begin{figure}
	\centering
	\includegraphics[width=0.5\textwidth]{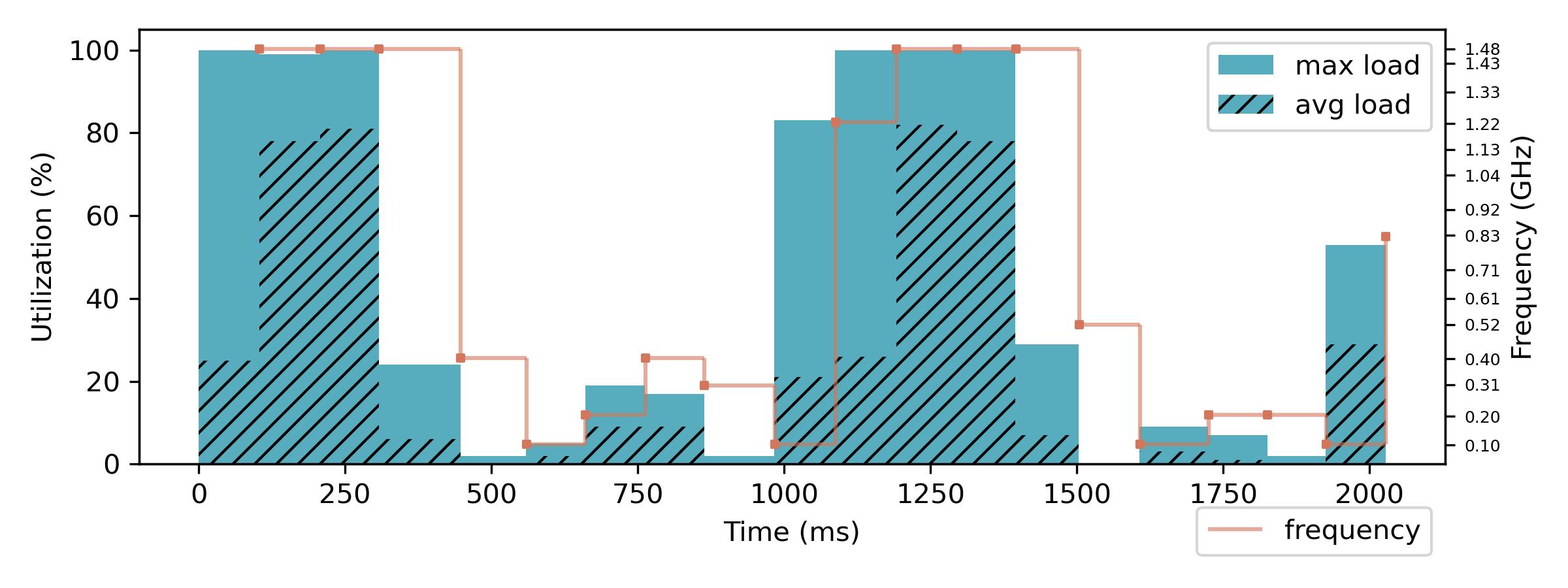}
	\caption{Profiling of \textit{Ondemand} governor using our profiler.}
	\label{fig:4}
\end{figure}

Fig.~\ref{fig:4} shows a visualization of \textit{Ondemand} governor's runtime profiled by the proposed method.
The system in this example runs a face recognition workload per second (the blue dash line in the figure is the \textcolor{black}{task period} and the deadline).
At each sampling point (orange nodes in Fig.~\ref{fig:4}), the profiler records the maximum/average CPU load (utilization) among cores in the last period, and the subsequent frequency \textit{Ondemand} governor decides to set.
The timestamp of each action is precisely recorded in an ms-view.

%% file: sections/motivation.tex
\section{Linux built-in methods: limitations}
\label{sec:Limitation}

\begin{algorithm}
\label{algo:ondemand}
\caption{\textit{Ondemand} DVFS governor in Linux V5.13: a simplified description}
\begin{algorithmic}[1]

\State $F$ denotes the provided frequency options.
\State $min_f/max_f$ denotes the min/max supported frequency in $F$.
\State $next_f$ denotes the frequency to be applied.

\For{each sampling period}
  \State Calculate the last period's CPU utilization $u$, $u \in [0, 1]$.
  
  \If {$u > up\_threshold$  (tunable, $\in [0, 1]$, 0.8 by default)}
    \State $next_f = max_f$.
  \Else
    \State $next_f = min_f + (max_f - min_f) \times u$.
  \EndIf
  
  \State $next_f = (1 - powersave\_bias$ (tunable, $\in [0, 1]$, 0 by default)$) \times next_f$.
  \State $next_f$ = the highest frequency below or at $next_f$ supported in $F$.
  \State Apply $next_f$.
 
\EndFor
\end{algorithmic}
\end{algorithm}

As of Version 5.13, Linux provides three dynamic DVFS policies \cite{cpufreqdoc}: \textit{ Ondemand, Conservative, and Schedutil}.
\textit{Ondemand} and \textit{Conservative} are time triggered. 
They use a timer to regularly sample data from the past period and control the frequency of the next period.
\textit{Schedutil}, implemented as part of the scheduler, does not rely on a timer but is actively woken up by the scheduler to tune the frequency. 
The core strategies for \textit{Ondemand} and \textit{Schedutil} are similar. Algorithm 1 gives a description of \textit{Ondemand} in Linux V 5.13.
This strategy is effective in its ability to reduce frequency/voltage in low-utilization periods (see 300 ms - 1000 ms in Fig.~\ref{fig:4} for example), thus saving dynamic and static energy consumption.
The conservative governor slowly changes the frequency at a fixed pace (different from line 9 in Algorithm 1), which is less responsive to changes in utilization.

Linux built-in policies are designed for general-purpose systems, and they are low overhead and do not need prior knowledge of the workload. Thus they are the current practice of DVFS.
Next, we study the limitation of Linux built-in DVFS for soft-deadline workloads. We achieve this by identifying a few system settings and workload patterns that the built-in governors can not effectively handle. This motivates the development of a reinforcement learning governor for energy saving that needs little prior knowledge of the tasks and machine.

\subsection{Coarse-Grained Voltage/Frequency Support}

\begin{figure}
	\centering
	\includegraphics[width=0.5\textwidth]{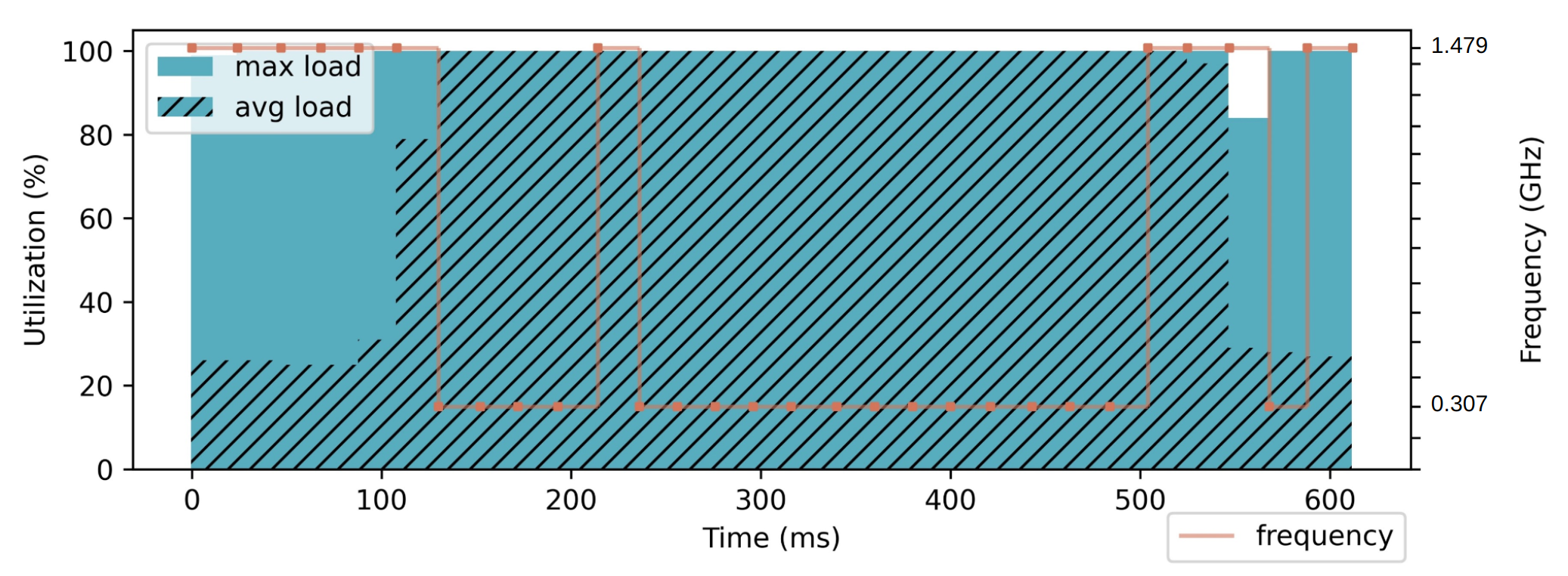}
	\caption{Use only two V/f supports to fill the fine slack.}
	\label{fig:10}
\end{figure}

\begin{figure}
	\centering
	\includegraphics[width=0.5\textwidth]{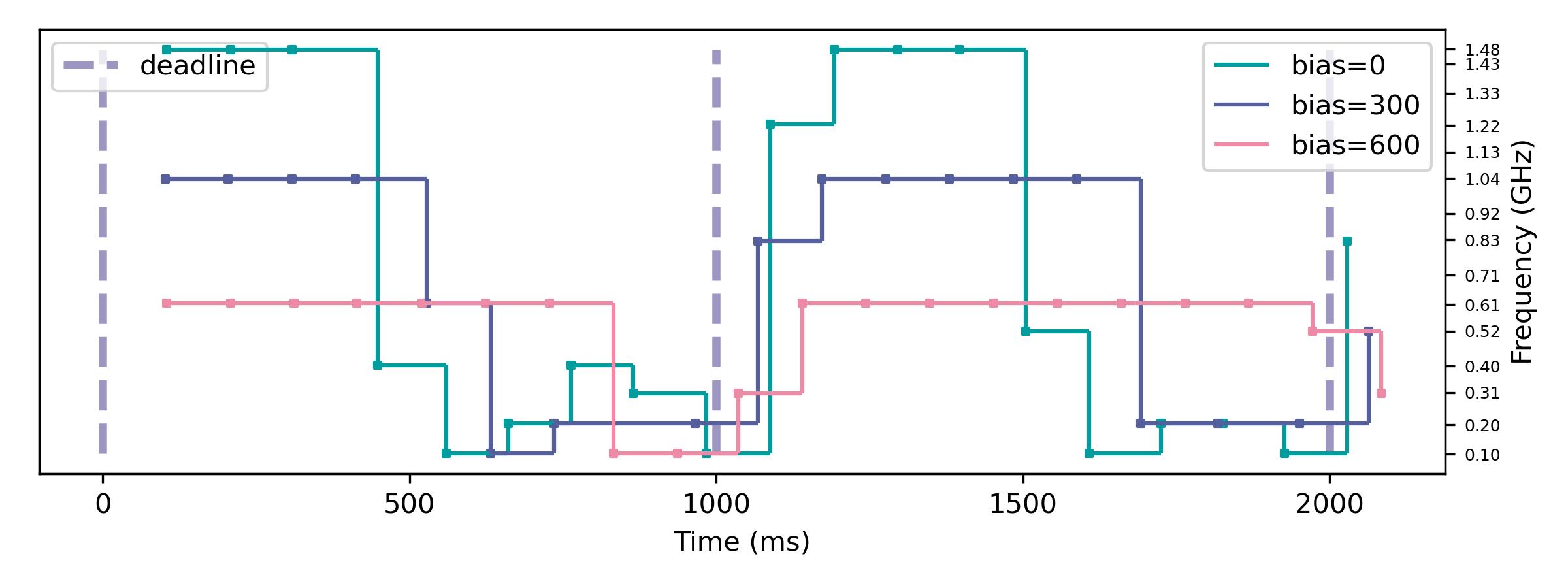}
	\caption{\textit{Ondemand} tuning when \textit{powersave\_bias} = 0, 0.3, 0.6}
	\label{fig:7}
\end{figure}

As Linux V 5.13, Linux has one built-in parameter  (\textit{powersave\_bias} in \textit{Ondemand}) for tuning DVFS (downscaling CPU performance to fill the slack).
The strategy is shown in line 10 of algorithm 1.
Fig.~\ref{fig:7} displays the tuning of \textit{powersave\_bias} on the workload shown in Fig.~\ref{fig:4}.

Such a strategy will be less effective for embedded CPUs that do not support fine-grained voltage/frequency support.
Although the ARM-A57 CPU in Nvidia Jetson Nano Board 2GB supports 15 frequency levels, it only supports two voltage levels, and energy-saving requires the CPU frequency to be reduced along with the voltage.
We can observe that the downscaling performed by the \textit{powersave\_bias} shown in Fig.~\ref{fig:7} cannot save energy.
% {\color{red} %not clear! 
Only when many frequencies are dropped to 0.307 GHz (a lower voltage value) the loss of performance begins to have energy-saving benefits.
This means a 5x CPU slowdown (drop from 1.479 GHz to 0.307 GHz in high utilization periods), and in many cases, the system would not have such a considerable slack to fill.
Conversely, if the system supports the fine-grained V/f option, the user can drop the CPU performance slightly to fill a small slack (e.g., from 1.479 GHz to 1.326 GHz).

As one of the most well-known embedded boards, the ARM-A72 in Raspberry Pi 4B also supports only 2 CPU V/f levels.
For Raspberry Pi 4B, users can edit \textit{/boot/config.txt} to enable the undervoltage function \cite{RBPDVFS}.
In this case, for the lowest frequency (0.6 GHz), the corresponding CPU voltage is reduced, but it also means a 2.5x CPU slowdown (from 1.5 GHz to 0.6 GHz).

In IoT systems where a large number of small devices need to be deployed, people tend to want cheaper devices and, therefore, potentially face challenges of energy savings for systems with coarse-grained V/f support. We want to point out that even Raspberry Pi 4 and the Nvidia Jetson Nano Board 2GB, the two relatively high-end embedded devices that nowadays cost more than a hundred dollars, support only coarse-grained V/f. Thus, coarse-grained V/f support is a common system setting that we need to consider when designing DVFS governors for embedded systems. 

\textbf{For machines that only support the coarse-grained V/f option, can the DVFS governor be tuned to fill the fine slack?}
For example, for the workload shown in Fig.~\ref{fig:4}, for the default setting of \textit{Ondemand} on Nvidia Jetson Nano Board 2GB, a task takes about 0.35 seconds to complete. %Is it possible to develop a strategy for the 0.6-second deadline?
Fig.~\ref{fig:10} shows such a possible solution for the 0.6-second deadline setting. However, such a solution cannot be found by tuning the built-in governors. If one wishes to tune an energy-saving policy based on \textit{powersave\_bias} for the built-in governor, the deadline for a task cannot be less than 1.5 seconds.

We observe that for the same high CPU utilization periods, the solution shown in Fig.~\ref{fig:10} sets part of the periods to high frequency and part of the periods to low frequency.
This is not possible for the built-in Linux methods since they determine the demand for the next period based on the utilization of the past period (the system's instantaneous computational demand).
In order to develop the strategy shown in Fig.~\ref{fig:10}, the governor needs to be able to develop the strategy without being bound to instant characteristics and based on the overall execution of a task.

\subsection{Unbalanced Load Distribution}

\begin{figure}
	\centering
	\includegraphics[width=0.5\textwidth]{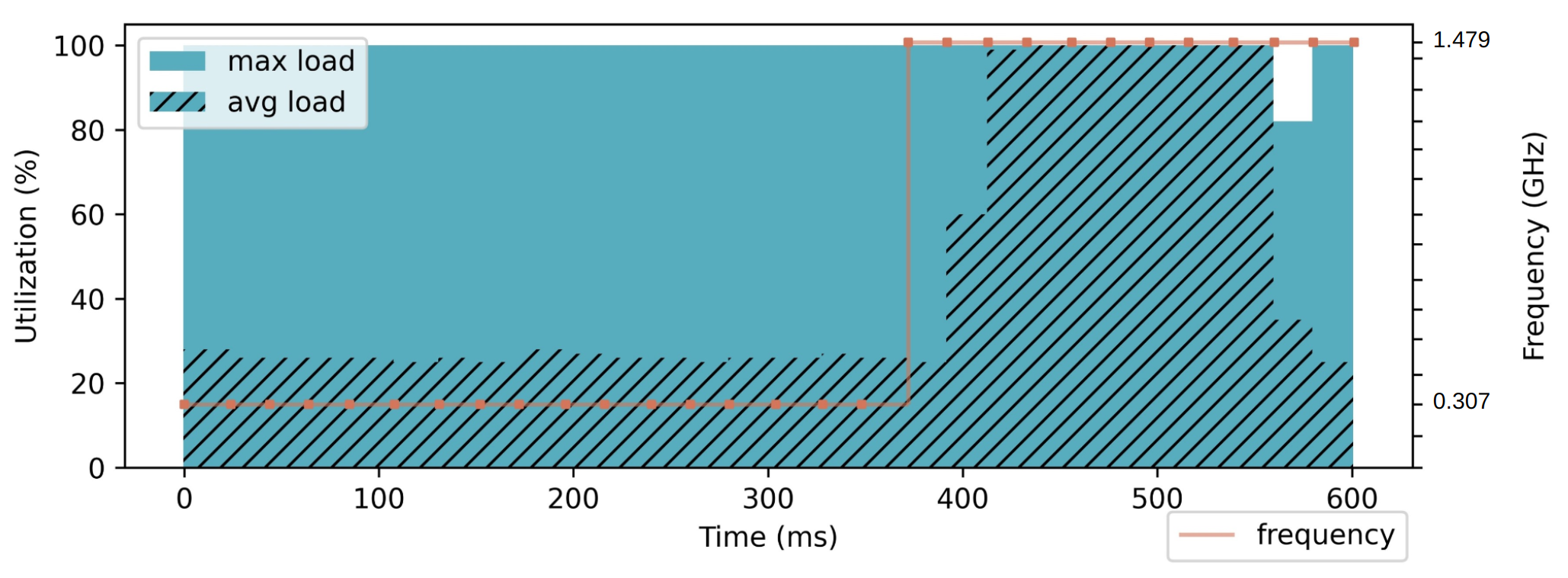}
	\caption{A strategy with lower average utilization.}
	\label{fig:9}
\end{figure}

\begin{figure}
	\centering
	\includegraphics[width=0.4\textwidth]{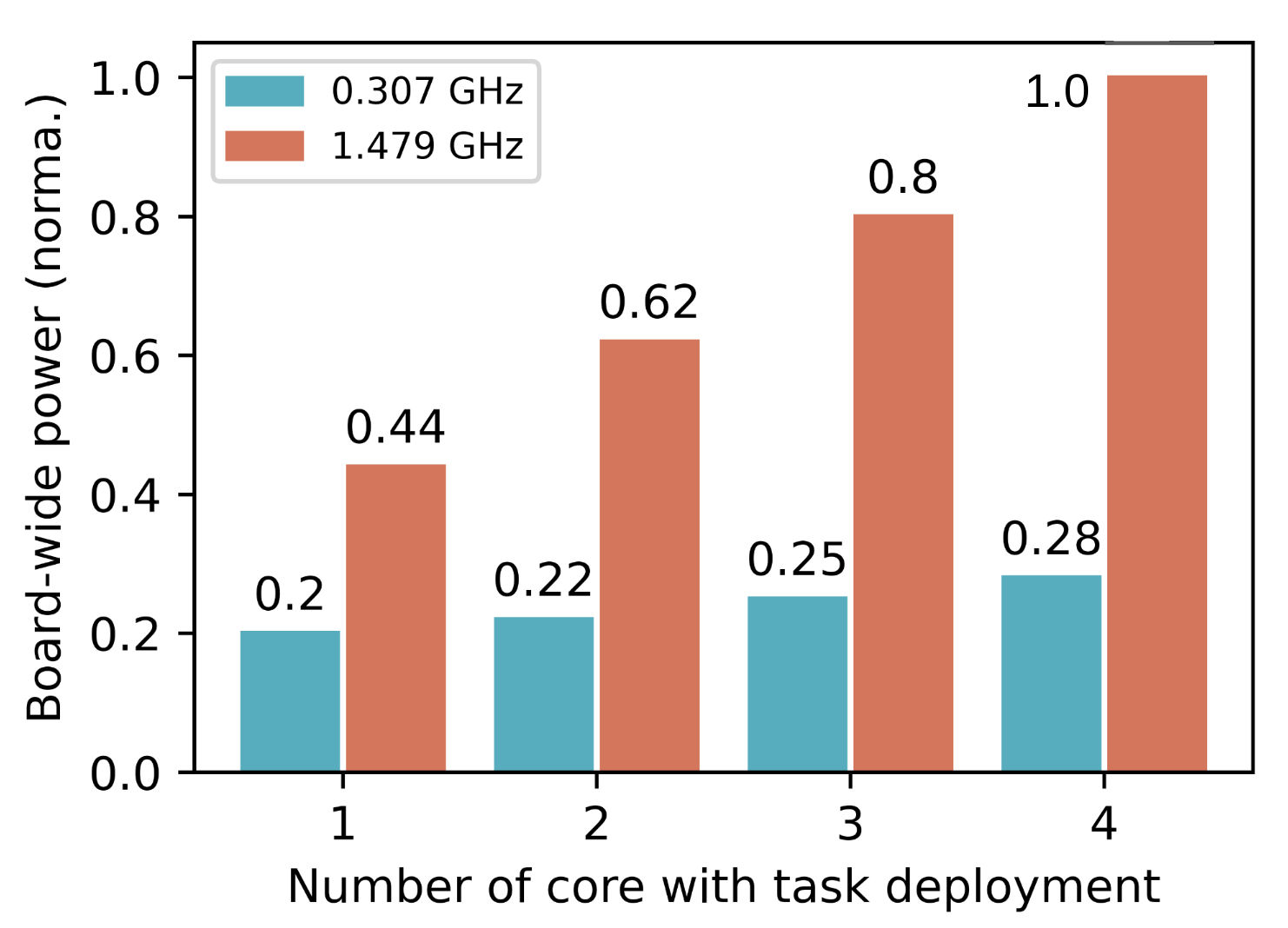}
	\caption{Power consumption for different CPU utilization on Jetson Nano Board 2GB.}
	\label{fig:e10}
\end{figure}

Since many embedded CPUs nowadays (such as the ARM-A57 in the Nvidia Jetson Nano Board 2GB and the ARM-A72 in the Raspberry Pi 4B) do not support per-core DVFS, the entire CPU package must run at one single frequency.
The Linux built-in method chooses the subsequent frequency based on the highest utilization among all the cores in the previous time period.

For periods with different average utilization, the same frequency downscaling may result in different energy gains with similar performance loss.
This is because a task will only be considered finished when all the tasks on all CPU cores have been completed.
For example, for the workload shown in Fig.~\ref{fig:4}, downscaling the period of 0-100 ms would have a similar performance loss as downscaling the period of 100-200 ms because the max CPU utilization among cores in these two periods is both 100\%.
In this case, executing the instructions in one period with 1.479 GHz can be converted to around five periods with 0.307 GHz.
However, downscaling the period of 100-200 ms could have more energy gains because of the higher average utilization (more cores at work), which means more dynamic energy consumption caused by 0/1 flipping can be saved.
As shown in Fig.~\ref{fig:e10}, downscaling V/f on higher-average-utilization periods can save more power, which means more energy-saving when the running time is consistent.

Fig.~\ref{fig:10} (denoted by $Policy_{high\_util}$) and Fig.~\ref{fig:9} (denoted by $Policy_{low\_util}$) show two DVFS strategies on workload \ref{fig:4} when the deadline for one task is 0.6 s.
$Policy_{high\_util}$ can save more energy compared to $Policy_{low\_util}$.

Both strategies allocate a similar amount of time for the CPU to run at low frequency/voltage (around 57\% at 0.307 GHz and around 43\%  at 1.479 GHz).
In this way, they consume a similar amount of static energy (formula \ref{ps}).
They differ in that $Policy_{high\_util}$ prioritizes frequency reduction for periods with high average utilization (high avg load), thus resulting in more dynamic energy consumption (formula \ref{cubic}).

We would like to trade the same performance loss for more energy benefits.
When uneven load distribution occurs, the DVFS governor should give preference to periods with high average utilization for downscaling with similar performance loss, which results in higher average CPU utilization.
This is not possible for algorithmic architectures similar to Linux's built-in DVFS approach, which performs inference based on short-time system characteristics.

\subsection{Internal Slack}

The slacks discussed in the above examples all appear after the task execution has finished. Due to the presence of IO blocks, slack can also occur during the execution of a task.

\begin{figure}
	\centering
	\includegraphics[width=0.5\textwidth]{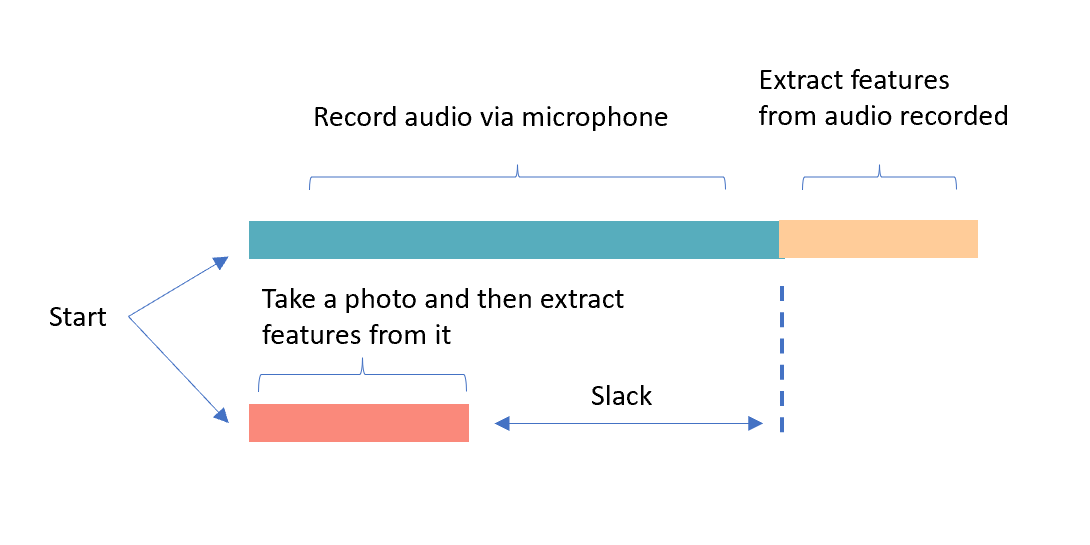}
	\caption{An internal slack example.}
	\label{fig:11}
\end{figure}

\begin{figure}
	\centering
	\includegraphics[width=0.5\textwidth]{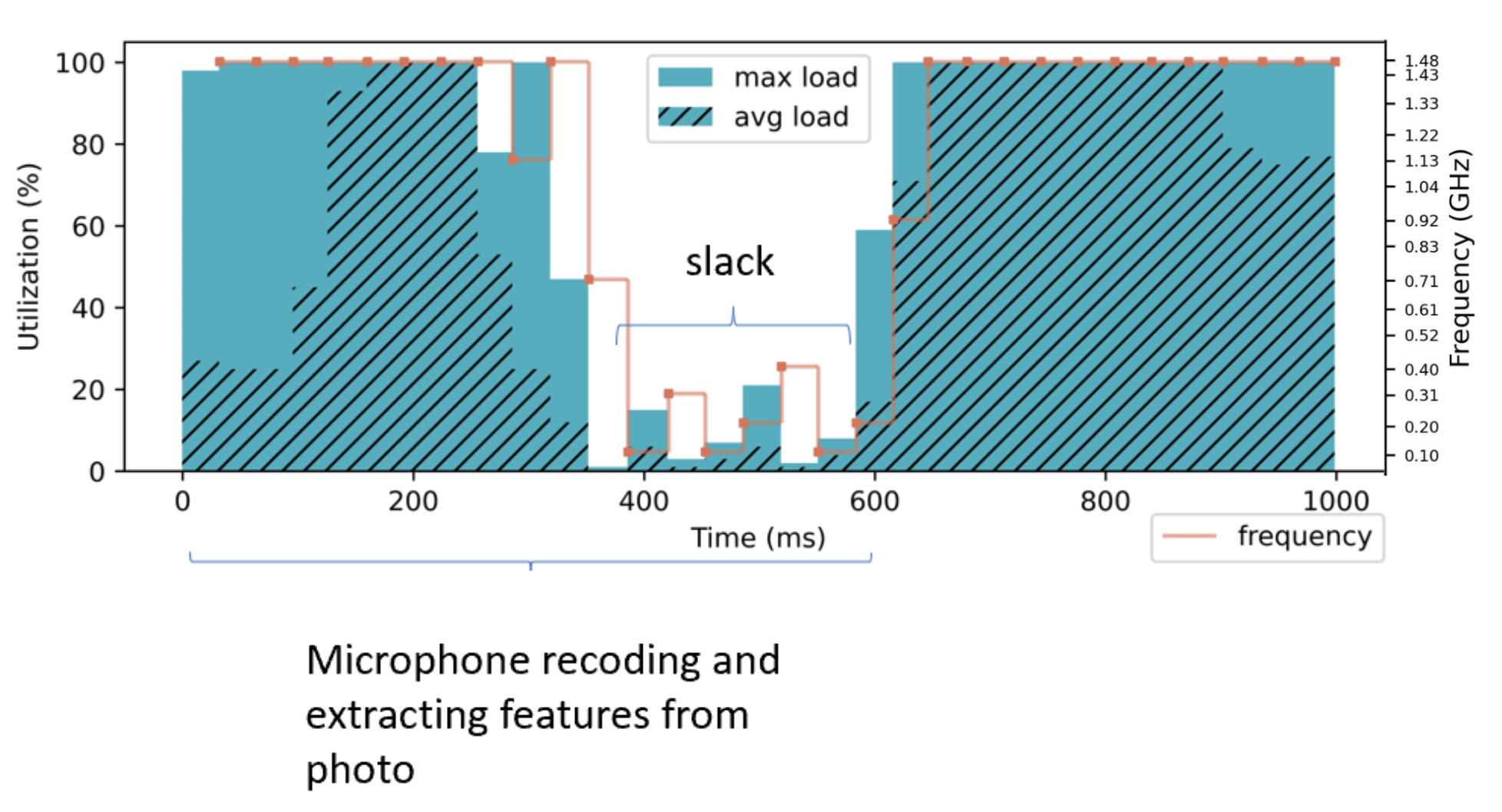}
	\caption{Profiling of default \textit{Ondemand} on the internal slack example.}
	\label{fig:12}
\end{figure}

Consider a scenario in which the system periodically performs feature analysis on both photos and audio.
The process of recording audio is typically an IO-intensive calculation, which can then be used to perform the photo analysis process.
The photo analysis process can be slowed down to fill the slack caused by recording audio, thus saving energy without compromising overall performance.
The photo analysis workload is still the face recognition program we used in previous sections. The microphone recording is set to be 0.6 s.
Fig.~\ref{fig:11} and Fig.~\ref{fig:12} show the pipeline of this example.

To handle this situation, the frequency governor should lower the frequency in the early stages and raise it in the later stages.
\begin{figure}
	\centering
	\includegraphics[width=0.5\textwidth]{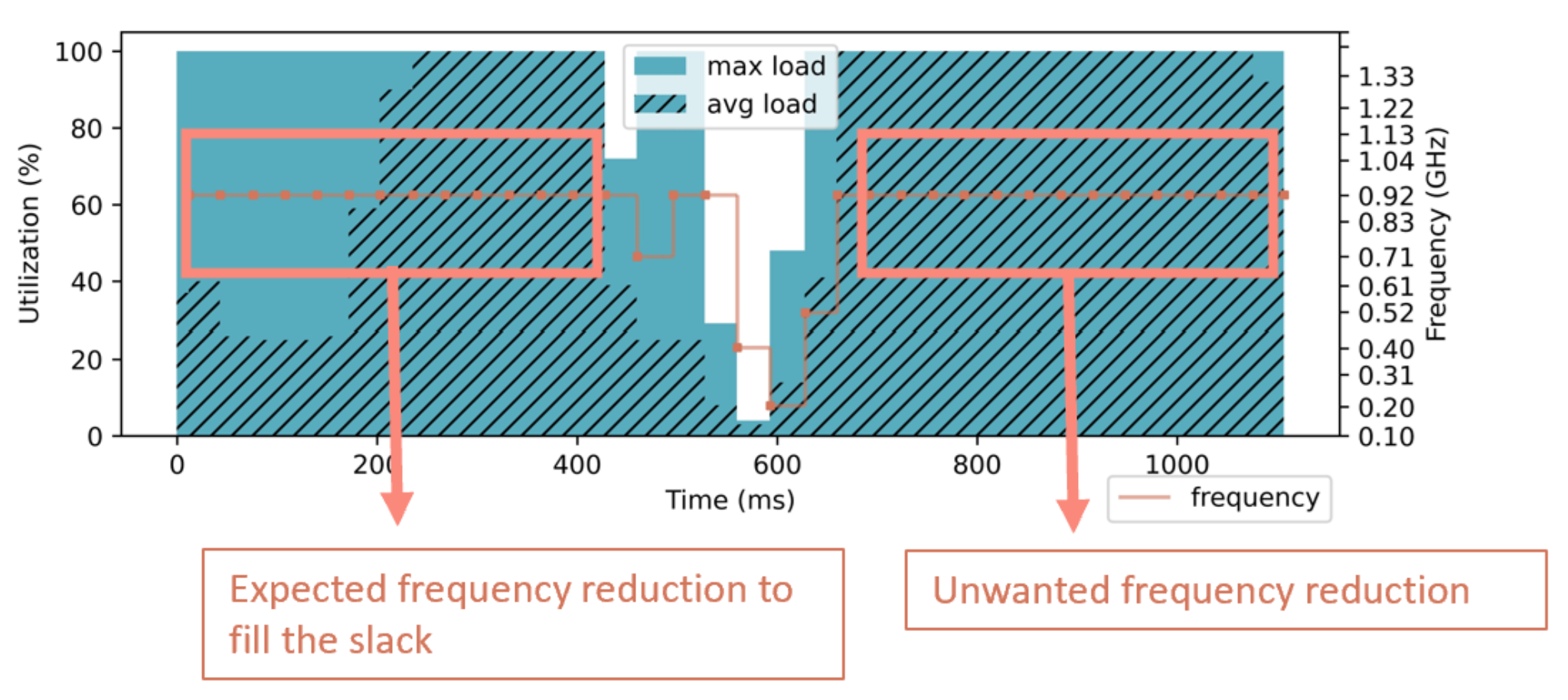}
	\caption{Fill the internal slack via \textit{Ondemand}'s \textit{powersave\_bias}.}
	\label{fig:13}
\end{figure}

If we use \textit{Ondemand} to fill this slack, there will be an inevitable performance loss (Fig.~\ref{fig:13}).
This is because, in the built-in governors' perspective, both the front and back sections are CPU-intensive computations, and they  should be treated equally.

\begin{figure}
	\centering
	\includegraphics[width=0.5\textwidth]{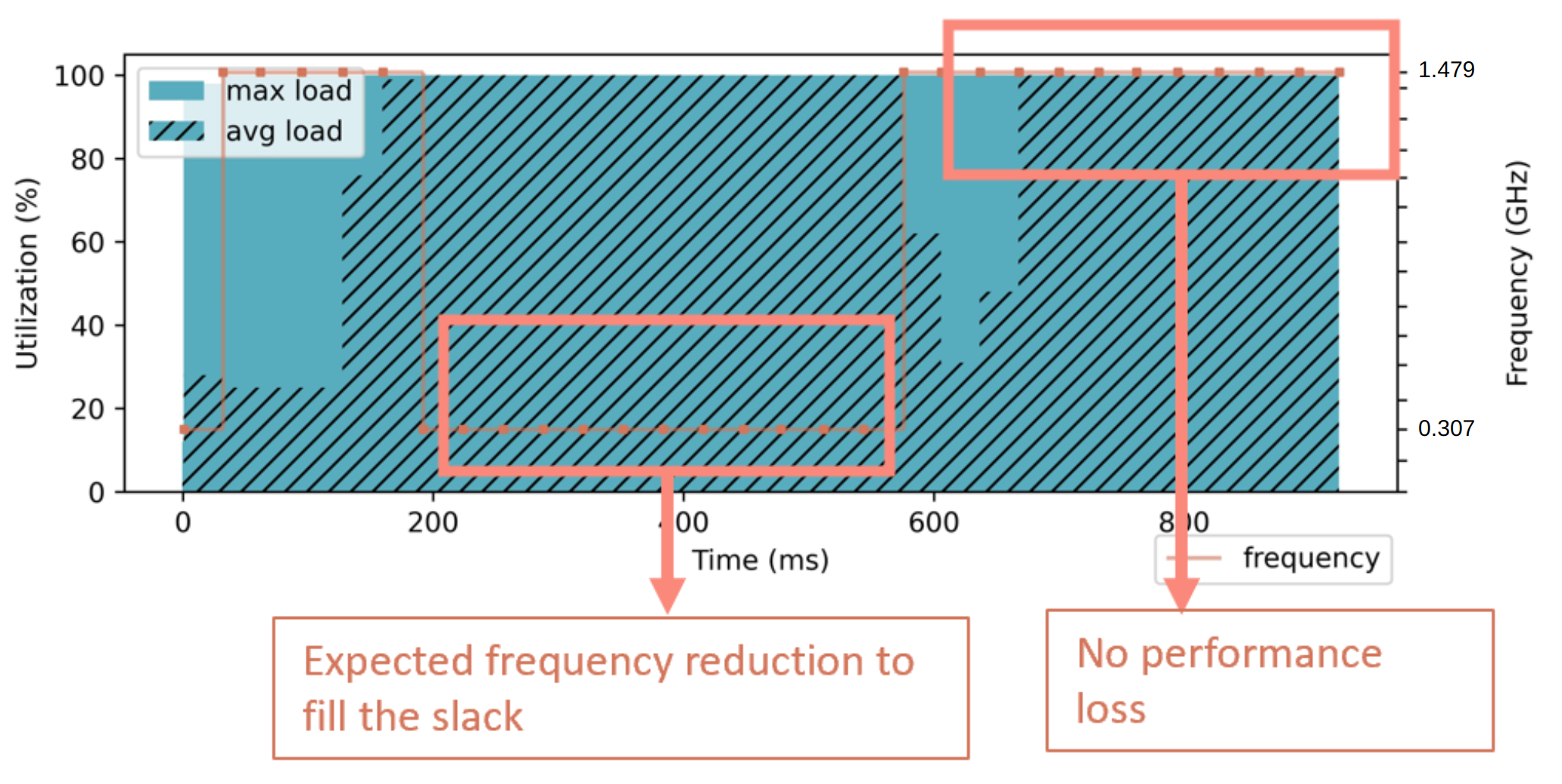}
	\caption{Fill the internal slack without performance loss.}
	\label{fig:14}
\end{figure}

In order to be able to fill the slack without losing performance, the governor needs to further understand the characteristics of the task.
Fig.~\ref{fig:14} shows such an example.

In fact, such internal slacks are common for IoT applications.
Modern IoT systems \cite{brunelli2019energy}\cite{taneja2019connected}\cite{chintanpalliiomt}\cite{togneri2019advancing}\cite{sahraoui2020deepdist} contains a variety of sensors and network calls.
Longer slacks include microphone recording, video recording, etc. Shorter slacks include temperature detection, etc.

\subsection{Why Extending this DVFS Framework cannot Cope with the three Patterns?}

% \textcolor{cyan}{
One CPU frequency control flow of the Linux built-in method can be summarized as follows.
\begin{enumerate}
    \item Collect system features for the past period. All three governors consider only CPU utilization in this step. \textit{Ondemand} and \textit{Conservative} use the calculation of CPU runtime divided by total time, and \textit{Schedutil} uses the PeLT metric provided by the scheduler.
    \item Predict the events of the next period. All three governors predict that the computational demand for a future period is consistent with that of a past period.
    \item Determine the CPU frequency for the next period using predicted events. Taking \textit{Ondemand} as an example, it selects the highest frequency if the utilization is above a threshold. Otherwise, it sets the frequency in equal proportion.
\end{enumerate}

The three patterns we discuss in this section are difficult to handle because the instance CPU utilization cannot capture the system state for making a good frequency decision for a given workload with a soft deadline. 
As shown in Fig.~\ref{fig:10} and Fig.~\ref{fig:14}, instances with the same measured utilization can be assigned different frequencies to reduce energy consumption most effectively.

% \textcolor{cyan}{
How can we make a DVFS governor aware of the difference between computing requirements?
An intuitive approach is to use more performance counters, making the system characteristics complex enough to distinguish.
But there are two problems with doing so.
\begin{enumerate}
    \item The usefulness of performance counters is specific to the workload. Adding extra performance counters may or may not help. The workload in Fig.~\ref{fig:14}  will benefit from using the counters provided by the microphone hardware, but that in Fig.~\ref{fig:10} will not benefit from using them. Whether the problem can be solved by adding more performance counters is case-by-case, and it is challenging to transfer a solution from one application to another.
    \item As the complexity of the inputs increases, it becomes more challenging to develop control strategies. The Linux built-in methods only use a value that logically ranges from 0 to 1 (CPU utilization) and designs some policies based on its explicit meaning. When multiple counters are enabled as input, the meaning of the input is no longer intuitive and even requires some degree of data mining. We will need more powerful models to handle the input, and the Linux kernel's resource constraints prevent it from supporting sophisticated mathematical models.
\end{enumerate}

% \textcolor{cyan}{
In summary, the DVFS algorithm that makes frequency scaling decision based
on the system features of the past period has difficulty
 coping with the three patterns we discussed. We need a
DVFS governor that can understand the global workload
computation demands.

%% file: sections/method.tex
\section{Proposed Method}

In this paper, we design and implement a DVFS governor that adapts to workload requirements to better address the three challenges mentioned above.
The proposed method, like the Linux built-in method, only requires the system to provide CPU utilization as input and contains two important components: \begin{enumerate}
    \item \textbf{Temporal encoder}: Construct a state based on the observed sequence of features to better understand the progress of task execution.
    \item \textbf{Reinforcement learning driven component}: Develop a frequency control strategy based on trial-and-error experience.
\end{enumerate}

In this section we will present our design, and the method of implementation, separately.

%\subsection{Motivation}
\subsection{Temporal Features of Workload and Learning}
\subsubsection{Understanding Workload in terms of Time}

We observe that deadline-constrained periodic workloads have a special feature. That is, the workload has a fixed deadline and periodicity with stable tasks to run, which means a task run can be viewed as an episode. This gives two inspirations.
\begin{enumerate}
    \item The events that will take place for each execution are similar.
    \item The  DVFS governor can predict the events that will occur in the future if it understands both the complete events to be experienced and the events that have already occurred.
\end{enumerate}
In addition, the CPU utilization and its corresponding frequency over a period of time reflect the number of 0/1 bits flipped by the CPU.
The CPU frequency, utilization, and period experienced by the CPU imply the progress of the task execution. For the DVFS governor, this is a set of observed sequences.
We want to mine the workload execution information implicitly contained in this sequence to help the DVFS governor better understand the task's requirements.

\begin{figure}
	\centering
	\includegraphics[width=0.5\textwidth]{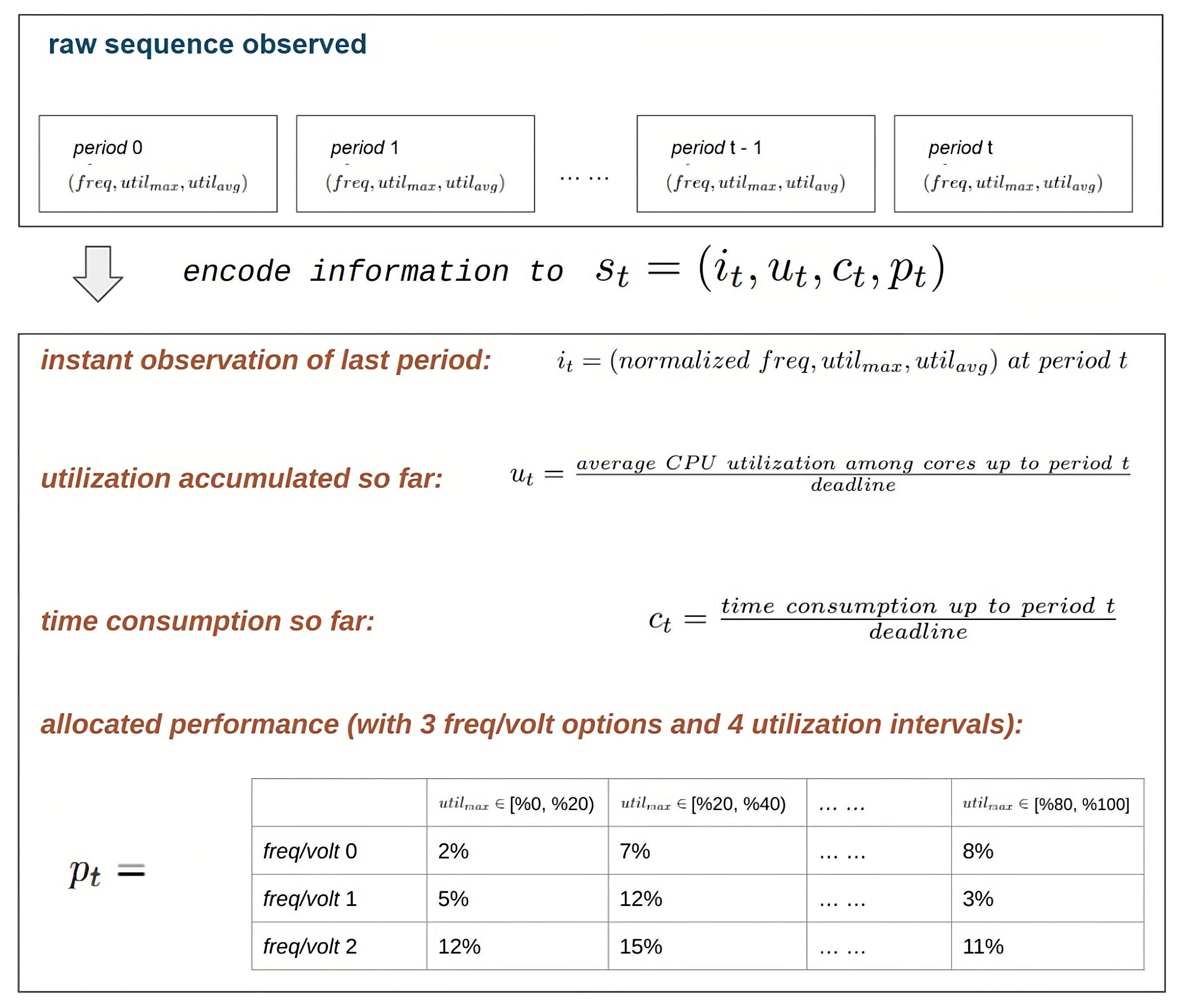}
	\caption{Encoding the observed time sequence to construct a state.}
	\label{fig:encoder}
\end{figure}

\subsubsection{Explicit Temporal Encoding}

We first define the time series observed by the DVFS governor.
For a periodic soft deadline real-time system, the system executes a task every \textit{T} seconds. \textit{T} is also used as the deadline of one task execution.
A DVFS governor, such as \textit{Ondemand} or \textit{Conservative}, performs frequency adjustment according to a pre-defined period.
For example, if a DVFS governor is set to adjust the CPU frequency ten times per second, it operates with a period of 0.1 s.
In practice, a DVFS governor cannot work strictly according to the set period. Some system events, such as hanging at idle moments, can affect the length of a period. Therefore, each period can be of variable length.

Without enabling additional performance counters for each period, the DVFS governor observes the CPU utilization and CPU frequency within that period.
This paper considers time series consisting of these observations.
Fig.~\ref{fig:encoder} (top portion) shows the format of a raw observed sequence.

The time series experienced by DVFS governor is indefinitely long. They contain intuitively useful information, but the question is how to understand the time series and develop strategies that produce time series resulting in low energy consumption. A previous work \cite{zhou2021deadline} used a Recurrent Neural Network (RNN) for the adaptive processing of time series.
However, this leads to lengthy training times, poor interpretability, and results in model architectures that are tuned to task needs.
While adaptive extraction of features is more in line with the definition of AI, doing so relies on powerful learning algorithms. 

In this work, we consider OS kernel-level code's inherent efficiency and reliability requirements and propose a method to develop a lightweight and interpretable learning and inference scheme. We extract information based on domain knowledge from an observed time series to provide a highly interpretable and low-dimensional encoding scheme.
The pipeline is shown in Fig.~\ref{fig:encoder}.

A time series at time $t$ is encoded as $s_{t} = \{ i_{t}, u_{t},  c_{t}, p_{t}\}$, shown in Fig.~\ref{fig:encoder}. Next, we explain each component.

$i_t$ denotes the observation of the past period. $i_t$ is to help the governor predict the current position of workload.
This information is also used by the built-in DVFS method of Linux.

\textcolor{black}{$u_t$ denotes the average CPU utilization up to sampling point $t$ in the current task period.}
$u_t$ is to help solve the problem of unbalanced load distribution that occurs in multi-core architectures. In the case of a similar performance impact of frequency tuning, priority is given to downscaling the periods of high average utilization, which saves more energy and shows an increase in the overall average utilization. 
This information is mainly intended to serve the purpose of exploring DVFS strategies based on reinforcement learning, which we will discuss in detail in the next section.

\textcolor{black}{$c_t$ denotes task progress up to sampling point $t$ within the current task period, ranging in [0, 1]}.
With the CPU performance allocation information, we introduce the time consumption progress $c_t = \frac{time\ consumption\ up\ to\ t^{th}\ period}{deadline}$.
We want the DVFS governor to be able to combine the use of $p_t$ (described below) and $c_t$ to understand the events that have been experienced.

$p_t$ is to help encode an abstract concept of "\textit{what events have happened and what events are to occur in the future}" and make it concrete into data that the DVFS governor can process.
In a period, the CPU's frequency reflects how fast it flips 0/1 bits, and the CPU's utilization and frequency reflect how much workload has been completed.
For a sequence, separate statistics on the usage of each CPU frequency in each utilization interval can give a guide to understanding how much workload has been completed.
The allocated performance matrix shown in Fig.~\ref{fig:encoder} gives an example, where we count 3 frequency/voltage usages with few utilization intervals.

\begin{algorithm}
\label{algo:encoder}
\caption{Temporal encoder}
\begin{algorithmic}[1]

\State \textbf{INPUT}: The observation of $t^{th}$ period: $(freq$, $util_{avg}$, $util_{max})$, and time consumption \textit{x} during this period;
\State  $s_{t-1} = \{i_{t-1}, u_{t-1}, c_{t-1}, p_{t-1}\}$, denoting the state for series from period $0$ to $t-1$

\State \textbf{OUTPUT}: $s_{t}$, denoting the state encoded from period \textit{0} to period \textit{t}.
\State $freq_{normalized} = \frac{freq - freq_{min}}{freq_{max} - freq_{min}}$.
\State $i_t = (freq_{normalized}$, $util_{avg}$, $util_{max})$.
\State \textcolor{black}{$u_t = u_{t-1} + \frac{ x \times util_{avg}}{deadline}$.}
%\State $u_t = \frac{u_{t-1} \times deadline + x \times util_{avg}}{deadline}$.
\State \textcolor{black}{$c_t = c_{t-1} + \frac{ x}{deadline}$.}
%\State $c_t = \frac{c_{t-1} \times deadline + x}{deadline}$.
\State $interval_{idx}$ = the index of the utilization interval $util_{max}$ belongs to.

\State $p_t = p_{t-1}$.
\State \textcolor{black}{$p_t[freq][interval_{idx}]= p_{t}[freq] [interval_{idx}] + \frac{x}{deadline}$.}
%\State $p_t[freq][interval_{idx}] = \frac{p_t[freq][interval_{idx}] \times deadline + x}{deadline}$.
\State $s_t = (i_t, u_t, c_t, p_t)$.

\end{algorithmic}
\end{algorithm}

\begin{figure*}
	\centering
	\includegraphics[width=1.0\textwidth]{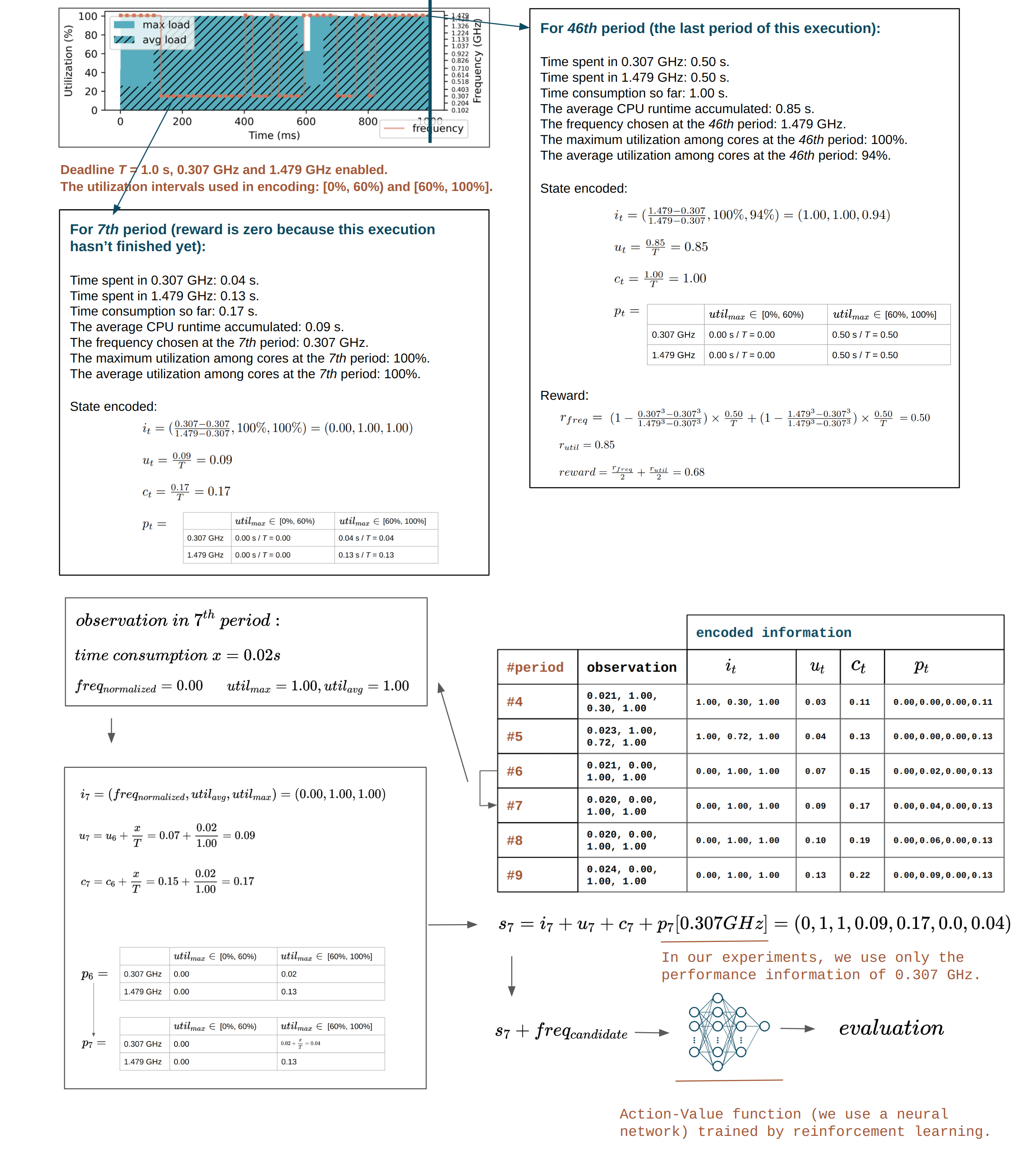}
	\caption{\textcolor{black}{Examples showing how states and rewards are calculated}.}
	\label{fig:example}
\end{figure*}

Overall, for a sequence, we encode its information explicitly into four parts: $u_t, i_t, c_t$, and $p_t$. The pseudo-code is shown in algorithm 2.
Fig. \ref{fig:example} shows examples of how we encode the observed temporal sequence into a state and how we assign a reward value to it. The purpose of the encoded information is to include the cause for insufficient or excessive task execution performance. We next describe how we use reinforcement learning to construct a DVFS policy based on this information. Ideally, we would like to use reinforcement learning to summarize which execution sequences lead to an encoding state with a high reward value (low power consumption and satisfying performance requirements), and the model can select actions during execution to bring the encoded state closer to 
a final state with a high reward value.

\subsubsection{Reinforcement Learning Driven Policy Development}

One challenge in considering complex features in the CPU frequency control process is how to map the information to the final decision.
The Linux strategy is to use features that are simple and explicitly contain useful information. For example, in \textit{Ondemand}, future frequencies are linearly equated to the observed utilization.
When the features under consideration become complex, it becomes more challenging to design a heuristic strategy.
We use reinforcement learning to summarize control strategies from experience.

\begin{figure}
	\centering
	\includegraphics[width=0.5\textwidth]{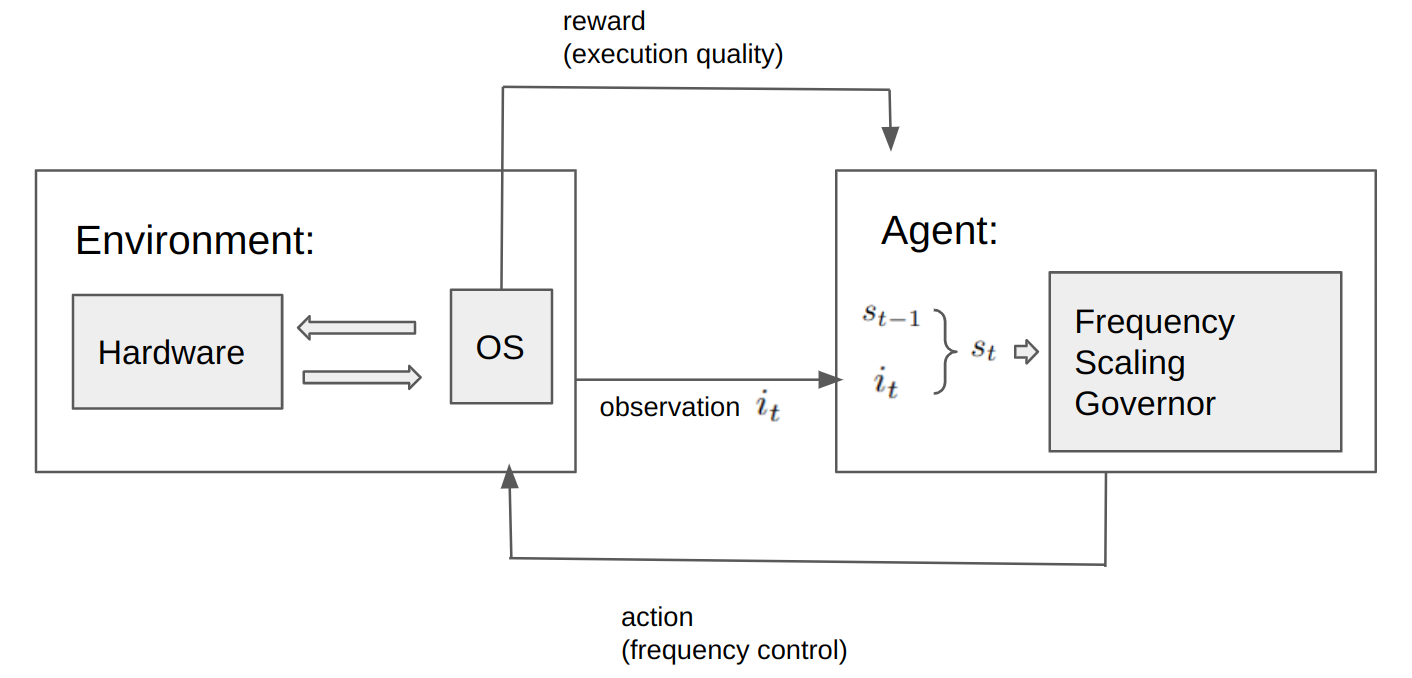}
	\caption{Frequency scaling as a reinforcement learning scenario.}
	\label{fig:rl}
\end{figure}

Reinforcement learning is a class of algorithms that observe the reward values harvested from behaviors and then explore strategies that can collect high reward values.
In reinforcement learning, one transition is defined as ($s_t$, $a_t$, $s_{t+1}$, $r_{t}$), where $s_{t}$ the current state, $a_t$ the action taken, $s_{t+1}$ the resulting state, and $r_{t}$ the immediate reward assigned to $(s_t, a_t)$.
A reinforcement learning algorithm uses such transitions to update its value-action function, which is used to evaluate the optimality of an action (the CPU frequency, in our case) for a given state.
Fig.~\ref{fig:rl} shows how to model CPU frequency scaling as a reinforcement learning scenario.

\begin{algorithm}
\label{algo:reward}
\caption{reward calculation for $s_t$}
\begin{algorithmic}[1]

\State Let $T$ denote the deadline for one execution.
\State Let $F$ denote the frequency table provided by hardware.
\State Let $f_{max}/f_{min}$ denote the max/min frequency supported in $F$.
% \State Let $P_f$ denote the power consumption under frequency $f$ with 100\% CPU utilization.
% \State Let $f_t$ denote the frequency selected at the $t^{th}$ step.

\If {$s_t$ is not the last state before the deadline reached}
    \State $r_{t} = 0$
\Else
    \If {\textcolor{black}{deadline missed}}
        \State $r_{t} = 0$
    \Else
        \State $x =  \frac{time\ spend\ in\ f\ during\ T}{T}$
        \State $r_{freq} = \sum\limits_{f \in F} (1 - \frac{f^3 - f_{min}^3}{f_{max}^3 - f_{min}^3}) \times x$
        \State $r_{util}$ = \textit{the average CPU utilization during T}
        \State $r_{t} = \frac{r_{freq}}{2} + \frac{r_{util}}{2}$
    \EndIf
\EndIf

\end{algorithmic}
\end{algorithm}

\textcolor{black}{We want the model to develop an ideal policy by harvesting more rewards for each workload. The principle of designing a reward at a state is to reward a state that leads to low energy consumption, high average utilization and satisfies deadline.}

Our reward definition has three components: a reward for low-frequency selection (considering energy), a reward for high CPU utilization, and a penalty for exceeding the deadline (too low performance).
The reward value is 0 when the deadline miss occurs. Otherwise, the reward $\in$ [0, 1].
All transitions after the deadline miss are discarded.
The calculation of the reward is shown in algorithm 2.

We design the reward to be sparse.
Only at the end of the last transition does the model receive the non-zero reward value. The model will only receive a value of 0 as an immediate reward value in all other transitions.
We design it this way for two reasons.
\begin{itemize}
  \item We want the model to predict a value $\in$ [0, 1], which reduces the possibility that the model parameters diverge for predicting large values.
  \item Without prior knowledge, it's hard to tell if a workload will be time out during its execution. In this case, if we want to assign an immediate reward, we can only confer rewards from the point of view of energy consumption for all the transitions before the deadline miss. We observe that this leads to a rapid tendency of the model to choose low frequencies, thus increasing the learning difficulty.
\end{itemize}

\textcolor{black}{
This reward value can be considered a heuristic energy consumption measure. The ideal way is to use the actual energy consumption, but measuring the energy consumption with
high accuracy in a short time requires special hardware
support. Therefore in this work, we use this heuristic measure.
}

For any reinforcement learning problem, there is a need for a model to learn an action-value function.
Array-based Q Learning \cite{watkins1989learning} is popular in system design for its easy implementation and low overhead.
However, since our state is continuous and large, %(in our experiments, a state is a seven-dimensional vector) space, 
we need a function approximator to reduce the memory footprint and speed up learning.

\textcolor{black}{
Specifically, in our scenario, the temporal encoder encodes the sequence as a vector of length six, and each element belongs to 0 to 1.
If we use table-based reinforcement learning (e.g., Q Learning), our first step is to discretize the state consisting of floating-point numbers so that it can be stored in a limited space.
This brings up two questions:
\begin{enumerate}
    \item The state space is likely to remain huge. Suppose we discretize each element of the vector to ten values, then the size of the entire state space is $10^6$, which is unaffordable in kernel space.
    \item It is not easy to judge whether a discrete solution is good enough. If the discrete method is too fine-grained (for example, discretizing each vector element to ten values), it will lead to high memory usage and low learning efficiency. If the discrete method is too coarse (for example, discretize each element of the vector to three values), it may result in the information contained in the state being ignored.
\end{enumerate}
}

\textcolor{black}{
It is worth mentioning that for standard table-based Q Learning, each state has its own Q Value, which means that learning based on one set of data cannot be applied to the knowledge of other states.
As an example, the information contained in states 96 and 97 may be close, but since their Q values are stored separately, states 96 and 97 are two completely different states in terms of model learning, and thus may lead to a decrease in learning efficiency.
Using a function approximator (e.g., a neural network) can improve these problems, as it can:
\begin{enumerate}
    \item Process floating-point numbers with a pre-defined memory size (the approximator's parameter size).
    \item Improve data utilization, since each update involves the parameters of the entire approximator (e.g., one backpropagation of the neural network). In this approach, an update to one state allows similar states to be updated as well.
\end{enumerate}
}

We use the Double Deep Q-Network (DDQN) model \cite{van2016deep}, which uses a neural network as the action-value function and a target network during training to reduce the overestimation of actions.

This method will eventually train a neural network that reads the state input and predicts the reward value that the candidate action will receive.

\begin{figure}
	\centering
	\includegraphics[width=0.4\textwidth]{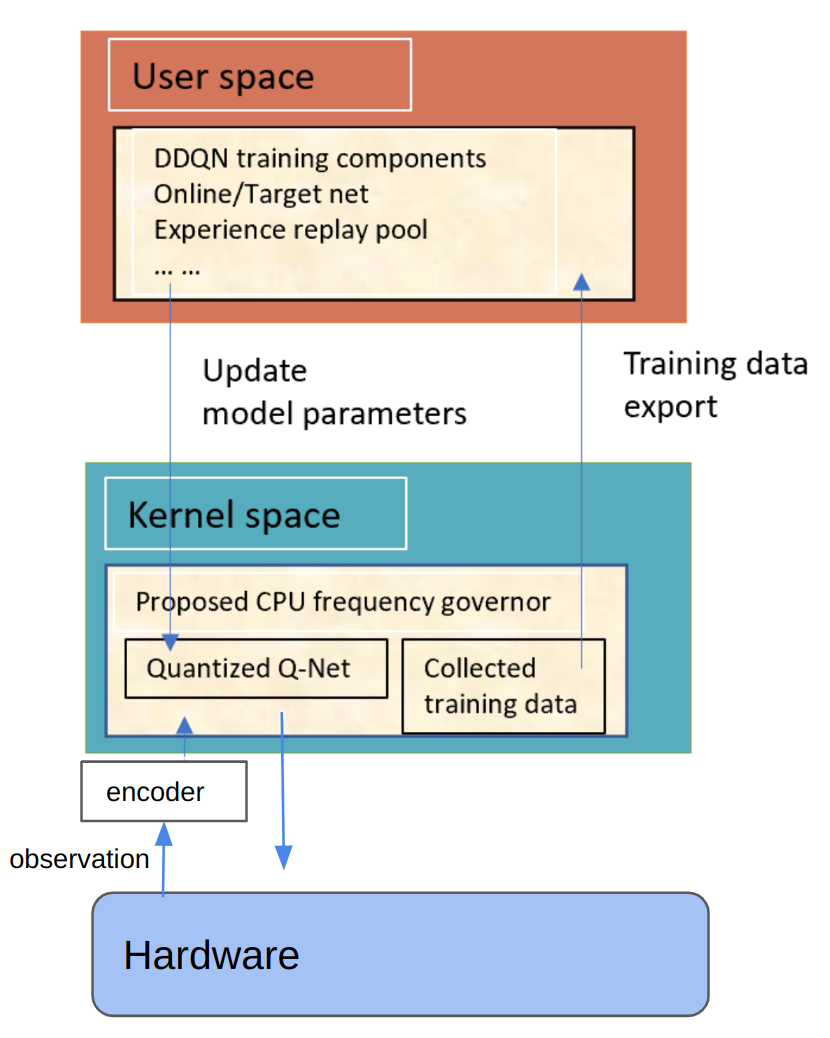}
	\caption{The user-kernel interleaving training framework with quantization.}
	\label{fig:16}
\end{figure}

\subsection{Implementation}

Similar to \cite{zhou2021deadline}, our work only implements the decision component at the kernel level. The learning component is implemented at the user level with data collected by the in-kernel profiler. Thus, the kernel state is only burdened with little inference overhead.

Our work further reduces the overhead by applying quantization \cite{jacob2018quantization}, with which the kernel can avoid floating-point calculation. As of today, the Linux kernel does not recommend the use of floating-point calculations. A complete integer-based kernel-state code would increase security (avoid breaking Linux design principles), enhance method pervasiveness (some low-end CPUs do not support floating-point computation), and reduce inference overhead on devices that are poorly optimized for floating-point computation.
The training pipeline is shown in Fig.~\ref{fig:16}.
 for states that are numerically close has been shown to be detrimental to the training of reinforcement learning models.

We used a simple quantization technique. 
With the state/reward design (all values are within [0, 1]), the parameters of the model and the values generated during calculation are within the range of [-10, 10]. We quantize all the values into [$-2^{30}$, $2^{30}$], and store them in 32-bit datatype (\textit{int} in C).
More advanced quantization techniques can further leverage the range of parameters and the datatype used, thus increasing the precision and reducing the memory footprint on huge-size models \cite{jacob2018quantization}. In our case, the model is small (around 150 parameters). We do not choose a more advanced quantization technique considering the expense of a more complicated code architecture needed for further optimization.

\textcolor{black}{In this work, we implemented our proposed governor in the kernel from scratch, including:
\begin{enumerate}
    \item A Linux C standard Neural Network engine, which can read the NN model resulting from the training by PyTorch in the user state and make inference based on this NN model inside the Linux kernel.
    \item An integer-based CPUFreq governor running under the CPUFreq framework that can be compiled as a Linux kernel module using the above engine for inference.
    \item A ring-buffer-based event profiler embedded inside CPUFreq governor. 
    \item A CPU frequency control visualization tool to visualize the information extracted by the above profiler.
    \item A set of protocols for the communications between the kernel space and the user space and control of the periodic workloads.
    \item A framework that assembles the above-mentioned modules to experiment with the proposed method in this work.
\end{enumerate}}
% Our approach to implementing NN-based inference on a device differs from other approaches like \cite{zhaoCAN2021} that generate Deep Learning C-code to run on a target device using the TVM \cite{TVM} compiler.
%Although many powerful tools exist for creating and training machine learning models in the user state, it is still challenging to deploy machine learning code in an underlying system environment with limited computational resources.
%Some intuitive methods include:
%\begin{itemize}
%    \item \textbf{Compiling a machine learning library into the target system}. The challenges of doing so include complex compilation rules due to the large amount of code to be added to the system's codebase, and the dependency of the computational library on other libraries. A simple example is that the Linux kernel does not even support the \textit{math.h} library in user-state C.
%    \item \textbf{Using advanced methods (for example, TVM \cite{TVM}) to generate code optimized on the target hardware \cite{zhaoCAN2021}}. The main reason we do not use this type of approach is that for our defined target requirements, we consider such an approach too complex for extensive experimentation.
%\end{itemize}
%}

\textcolor{black}{
With the interactive training-inference framework Fig. \ref{fig:16}, the amount of code added to the kernel internals is reduced, and there is no training overhead at the kernel level.
We implement a quantization-enabled neural network engine to read the model parameters generated by user-state PyTorch training and make inferences in the kernel.
The engine has a lean amount of code and is easy to compile into the Linux kernel. 
The codebase can be found at: \href{https://github.com/coladog/tinyagent}{https://github.com/coladog/tinyagent}.}
\textcolor{black}{Please refer to the project documents for more details of how we generate models from PyTorch and make inferences inside the Linux kernel, as well as the implementation of the quantization technique.}

\subsection{Training}

\definecolor{myred}{rgb}{0.77, 0.34, 0.33}
\definecolor{myblue}{rgb}{0.49, 0.61, 0.73}
\definecolor{mypink}{rgb}{0.99, 0.69, 0.67}

\begin{algorithm}[t]
\label{algo:kerneltrain}
\caption{Kernel-state inference module}
\begin{algorithmic}[1]

\State Initialize action-value function $Q$ with parameters $\theta$ trained in user space.
\State Initialize observed sequence $\phi = [\ ]$.

\For{each sampling period}

\State Calculate last period's observation: $(freq$, $util_{avg}$, $util_{max})$, and \textcolor{black}{the time spent during this sampling period.}
\State Encode state $s$ according to Algorithm 2.
% \State \textit{E}.add\_one\_period($freq$, $util_{avg}$, $util_{max}$, \textit{t}).
% \State $freq_{next}$ = M.inference().
% \State State $s$ = \textit{E}.construct\_state().
\State $freq_{next} = max_{action\_freq}Q^*(s, action\_freq; \theta)$.
\If {training required}
    \State Generate one random number $v \in [0, 1]$.
    \If {$v > \epsilon$}
        \State $freq_{next}$ = a random frequency.
    \EndIf
\EndIf
\State Apply $freq_{next}$.
\State Add ($s$, $freq_{next}$) into $\phi$.

\EndFor

\If {training required}
    \State Output $\phi$ into user space.
\EndIf

% \State Output $\phi$ to $B$.

\end{algorithmic}
\end{algorithm}

\begin{algorithm}
\label{algo:usertrain}
\caption{User-state training module}
\begin{algorithmic}[1]

\State Load action-value function \textit{Q} with parameter $\theta$. 
\State Initialize target function parameter $\theta' = \theta$.
\State Load replay memory \textit{D}.
\State Fetch newest training data $\phi$ output by kernel space, convert each node into a transition $(s_t, a_t, r_t, s_{t+1})$, and then add it into \textit{D}.
\State Construct training pool \textit{B} from \textit{D}.
\For{each batch $(s_t, a_t, r_t, s_{t+1})$ in \textit{B}}
\If{$s_{t+1}$ is non-terminal}
    \State $y_t$ = $r_t + \gamma Q(s_{t+1}, max_{a}Q(s_{t+1}, a; \theta); \theta')$.
\Else
    \State $y_t = r_t$.
\EndIf
\State Perform a gradient descent step on $(y_t - Q(s_t, a_t; \theta))^2$.
\State Every \textit{C} steps reset $\theta' = \theta$.
\EndFor
\State Output quantized $\theta$ into kernel space.

\end{algorithmic}
\end{algorithm}

\textcolor{black}{
The kernel state module observes new data at each step, encodes it into the current system state through the temporal encoder, and then uses Q Net to determine which action (frequency) will result in a larger reward.
Meanwhile, in the training phase, it randomly selects actions to explore different strategies according to certain odds and sends the observed sequences to the user state module.
The module in the user state collects the sequence data provided by the kernel state, calculates its corresponding reward value, and then uses a reinforcement learning algorithm to train the Q Net and return the updated parameters to the kernel state.
Through this interactive step, we finally implant a governor in the kernel state that understands workload requirements and saves energy while satisfying performance.
}

Algorithm 4 and algorithm 5 describe the training process using the interleaving framework. 
In the kernel state, the CPUFreq framework initializes an integer-based neural network (Q Net), reads the model parameters derived from the user state, and infers frequency actions based on this network at runtime.
In addition, the kernel state module records the data observed during runtime and exports it to the user state for training at the end of the run.
The user-state training module updates the model parameters according to the Double Deep Q-Network (DDQN) training method after loading the training data and model parameters.

In the process of generating the training pool (line 5 in Algorithm 5), we use the idea of prioritized experience replay \cite{schaul2015prioritized}.
The native experience replay pool considers each experience to be of equal priority, while the prioritized experience replay pool \cite{schaul2015prioritized} considers some data to be more worthy of learning, and thus purposefully selects some high-priority experience when constructing training data.

We divide the experience pool into 10 buckets and add a sequence of transitions to the corresponding bucket according to the reward received.
For example, when we have a sequence with reward = 0.27, then the sequence will be added to the $3^{rd}$ bucket, which contains the sequences with reward $\in [0.2, 0.3)$.
For each training pool construction, we will randomly take out 64 sets of sequences from each bucket and construct a training dataset using the transitions contained.
With this approach, the model can learn a variety of experiences with different reward levels in one training step.

%% file: sections/experimentation.tex
\section{Experimentation}

Learning an embedded control algorithm (a frequency control governor) for operating system kernels has not been widely explored by the industry or open-source community. Therefore there is a lack of open-source support. Our work has to build tools from scratch for efficient profiling, in-kernel inference, and frequency control policy visualization.

\subsection{Experimental Setup}

\begin{table}\centering
\label{table:setting}
\caption{Experiment settings}
\begin{tabular}{ |p{4cm}||p{4cm}|  }
% 	\hline
% 	Target Net Iter Frequency &   \\
	\hline
	Hardware & Nvidia Jetson Nano 2GB Board \\
	\hline
	OS & Linux 4.9 \\
	\hline
	Energy measurement & Board-wide energy consumption measured by a power meter \\
% 	\hline
% 	$\sigma$ in table 6.2 & 20 \\
% 	\hline
% 	$\varphi$ in table 6.2 & 64 \\
	\hline
	Optimizer & Adam with 0.001 learning rate \\
	\hline
	Batch size & 16 \\
	\hline
	Target Net updating frequency (\textit{C} in algorithm 5) & per 32 batch learning \\
	\hline
	NN structure & 7-8-8-1 with ReLU activation function \\
	\hline
	Utilization intervals used in encoding & \{[0\%, 60\%], (60\%, 100\%]\} \\
	\hline
	Sampling rate & 20000 ms, 50 times per second \\
% 	\hline
% 	Training episodes & 300 \\
	\hline
	Action randomly pick rate during training & 0.7 at the first 50 episodes, 0.5 at the next 50 episodes, and then 0.3 till the end \\
	\hline
\end{tabular}
\end{table}

We first verify whether our approach can better address the patterns we discussed in Section~\ref{sec:Limitation}. Two complex self-designed workloads are used for this purpose.
\begin{enumerate}
  \item \textit{FaceRecog}: The system periodically reads a photo and then identifies the faces' location. Image reading and pre-processing are done in a single thread, and face recognition is done in multiple threads. Fig.~\ref{fig:4} and Fig.~\ref{fig:7} show the frequency tuning of this workload under \textit{Ondemand}. The image processing is implemented based on OpenCV \cite{opencv_library}. 
  \item \textit{AudioRecog}: The system periodically performs an audio recording while running the \textit{FaceRecog} workload and performs feature extraction on an audio clip after the recording is finished. The audio analysis is implemented based on PyAudioAnalysis  \cite{giannakopoulos2015pyaudioanalysis}. Fig.~\ref{fig:11} and Fig.~\ref{fig:12} show the pipeline and the frequency tuning of this workload under \textit{Ondemand}.

\end{enumerate}

\begin{table}\centering
\label{table:setting}
\caption{\textcolor{black}{Workloads used}}

\begin{tabular}{ |p{4cm}||p{4cm}|  }
	\hline
	\textit{FaceRecog} & Self-constructed, including unbalanced load \\
	\hline
	\textit{AudioRecog} & Self-constructed, including unbalanced load and internal slack \\
	\hline
        \textit{Mibench} & Four workloads are used: \textit{bitcount}, \textit{susan}, \textit{dijkstra} and \textit{typeset} \\
        \hline
\end{tabular}
\end{table}

We also validate our approach on publicly available workloads with hidden implementation details.
In this step, we use four workloads provided by \textit{MiBench} \cite{guthaus2001mibench}: \textit{bitcount}, \textit{susan}, \textit{dijkstra} and \textit{typeset}.

On \textit{FaceRecog}, we train three sets of models corresponding to 0.6 s, 0.9 s, and 1.2 s deadlines, respectively.
On \textit{AudioRecog}, we train three sets of models corresponding to 1.0 s, 1.3 s, and 1.6 s deadlines, and with 0.6 s, 0.9 s, and 1.2 s microphone recording time, respectively.

We compare our approach with all DVFS governors (\textit{Performance}, \textit{Ondemand}, \textit{Schedutil}, \textit{Conservative}) currently supported by Linux, except for \textit{Powersave}, which just sets the frequency to the lowest level and thus can only be used in the extreme case.

Our experimental environment is the Nvidia Jetson Nano Board.
Our method uses two frequency options (1.479 GHz and 0.307 GHz) with different voltages.
However, more frequency levels must be enabled for Linux built-in methods, even with the same voltage support.
For \textit{Ondemand}, after calculating the logical frequency based on $min_f + (max_f - min_f) \times utilization$, it will lookup for a frequency supported by hardware that is below or at the logical frequency. In this case, if it only has two frequency support, unless the utilization is 0, it will always choose the higher one. 
\textit{Conservative} and \textit{Schedutil} also require fine-grained frequency support for similar reasons.
Therefore, we enable full frequency support for the Linux built-in methods, even though the hardware supports only two voltage levels.
\subsection{Reward Curve}

\begin{figure}
	\centering
	\includegraphics[width=0.5\textwidth]{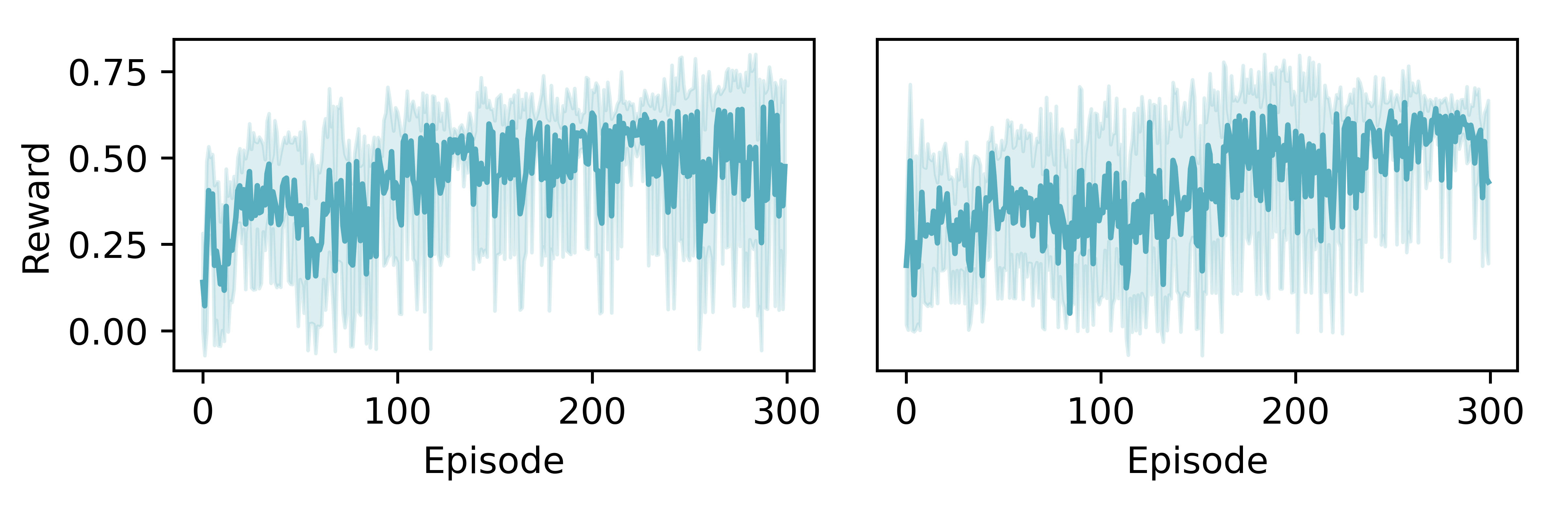}
	\caption{Reward curve with five training on \textit{FaceRecog} (left) and \textit{AudioRecog} (right).}
	\label{fig:e1}
\end{figure}

\begin{figure}
	\centering
	\includegraphics[width=0.5\textwidth]{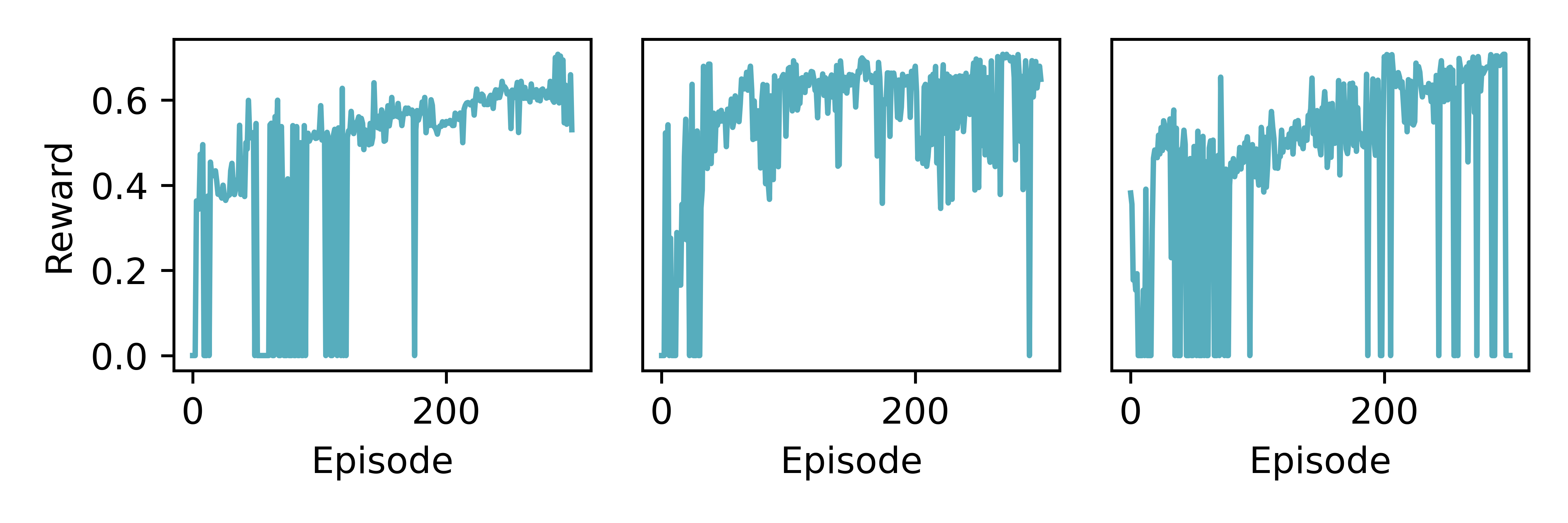}
	\caption{Three Reward curves with single training on \textit{FaceRecog}.}
	\label{fig:e2}
\end{figure}

After each training, we executed the workload five times using the latest model and then counted its average harvested reward value.
Fig.~\ref{fig:e2} shows three separate training on \textit{FaceRecog} with 0.6 s deadline.
Since training for reinforcement learning is subject to randomness (random selection of actions to explore, random learning of data), a common measure of learning quality is to take the reward curve of multiple learning sessions and count their mean and standard deviation.
Fig.~\ref{fig:e1} shows the result on \textit{FaceRecog} and \textit{AudioRecog}, with 0.6 s deadline and 1.0 s deadline separately.
Our method demonstrated the ability to harvest more reward value in training.

\subsection{Deadline Awareness}

\begin{figure}
	\centering
	\includegraphics[width=0.5\textwidth]{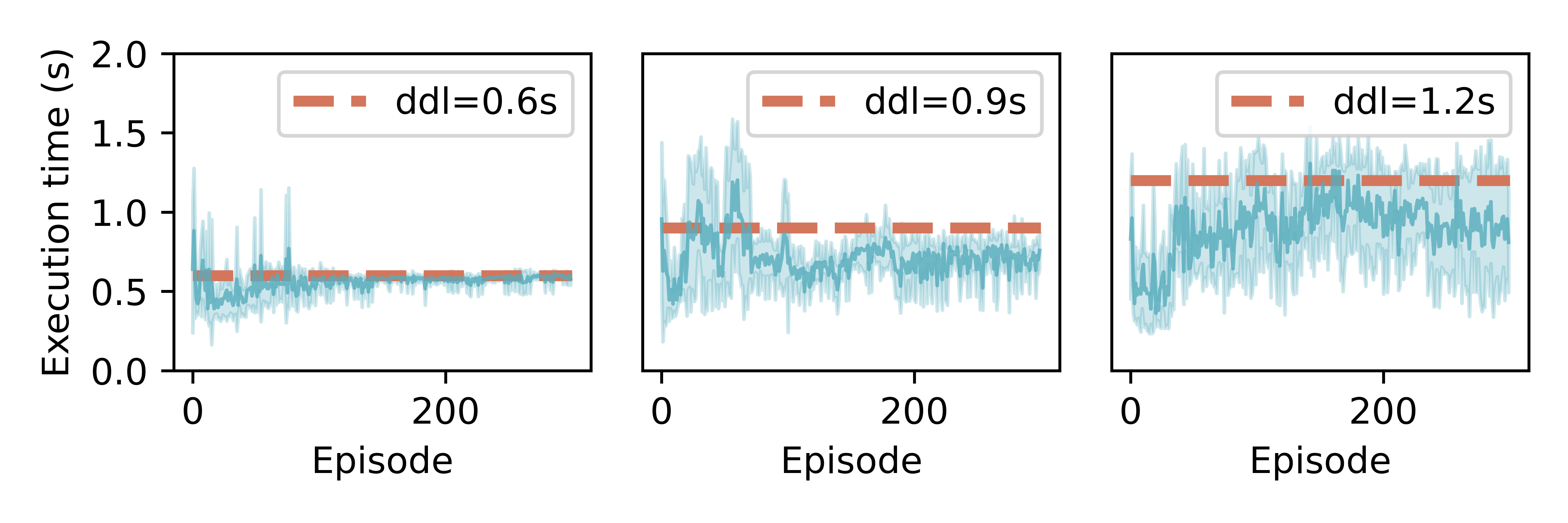}
	\caption{Execution time curve with five training on \textit{FaceRecog}.}
	\label{fig:e3}
\end{figure}

\begin{figure}
	\centering
	\includegraphics[width=0.5\textwidth]{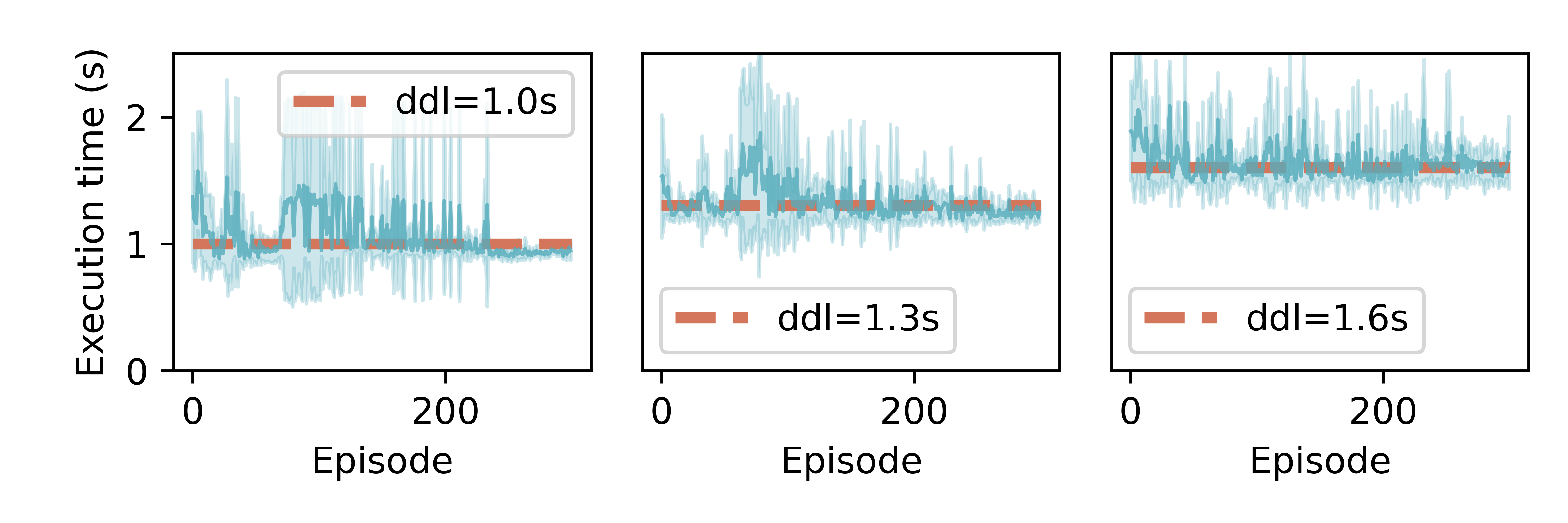}
	\caption{Execution time curve with five training on \textit{AudioRecog}.}
	\label{fig:e7}
\end{figure}

We also evaluate the task execution time under the policy after each training step five times and take the average.
Fig.~\ref{fig:e3} and Fig.~\ref{fig:e7} show the mean and standard deviation of the execution time curves based on five training sessions.
Our method perceived the need for deadlines very well.
\textbf{It is worth mentioning that for each deadline, our method receives only one numerical value (for example, 0.6 for 0.6 s deadline), but can generate a DVFS strategy accordingly that respects the deadline.}
In our experimental setting, the governor uses only two frequency values. 
%(1.479 GHz and 0.307 GHz) 
It cannot drop the overall frequency a little to accommodate the performance change (e.g., from 1.479 GHz to 1.428 GHz).
Adaptation to deadline requires it to combine the only two frequency options available.
It takes it upon itself to relate this abstract value to performance requirements and make policy adjustments.

\subsection{Learned Policy}

\begin{figure}
	\centering
	\includegraphics[width=0.4\textwidth]{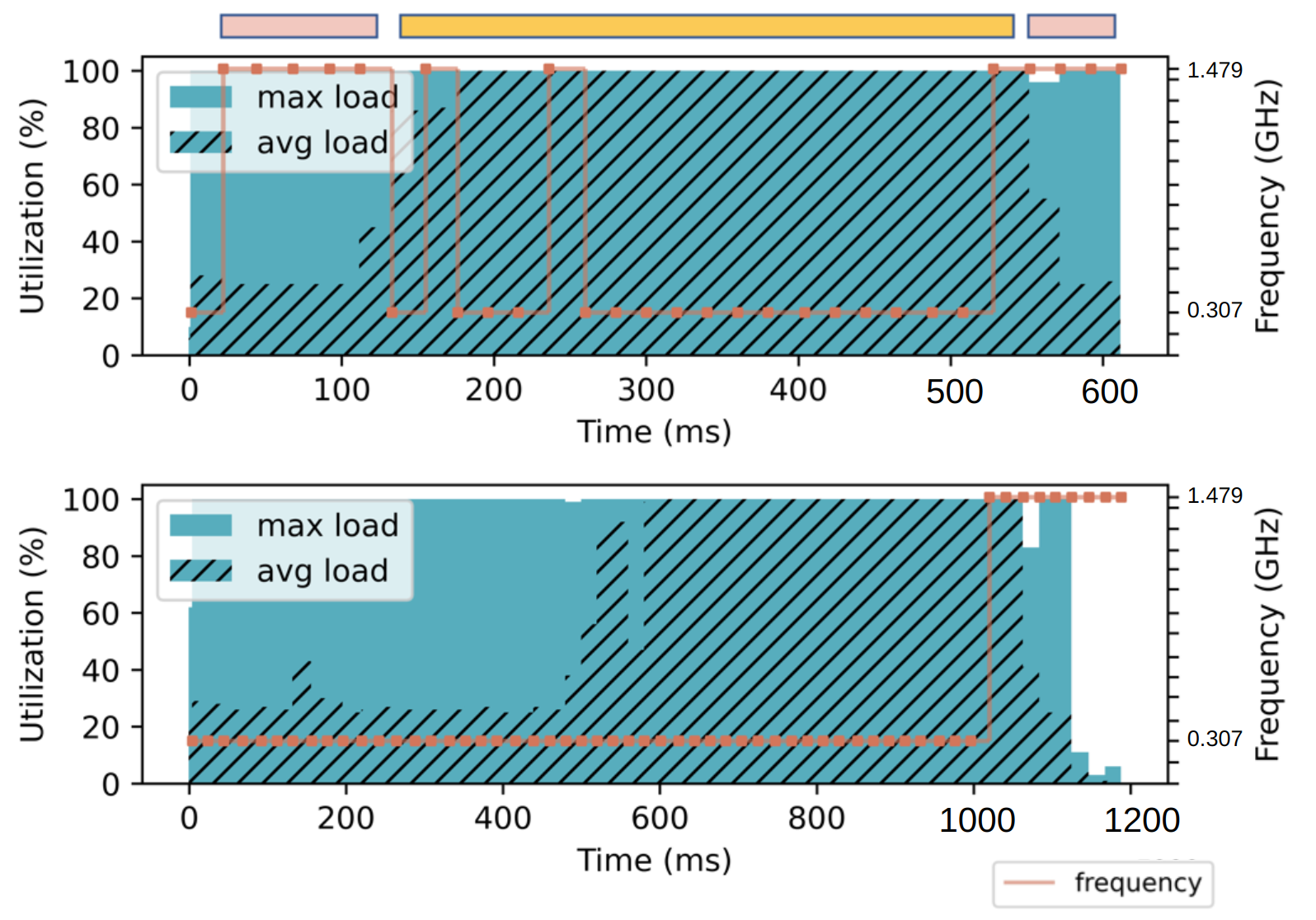}
	\caption{Frequency policy developed by proposed method on \textit{FaceRecog} with 0.6 s (top) and 1.2 s (bottom) deadline respectively.}
	\label{fig:e4}
\end{figure}

An important contribution of this work is visualizing the CPU frequency control process within the kernel for any workload with any given strategy. The visualization helps us understand and compare the differences between the policies and the learning process.

Through visualization, we observed that the proposed approach smartly learned very different strategies with the same workload when the deadline changed.
Fig.~\ref{fig:e4} shows an example. 
For different deadlines, although the maximum utilization among cores was close to 100\% throughout the execution, low frequencies were set to save energy without exceeding the deadline. 

\begin{figure}
	\centering
	\includegraphics[width=0.4\textwidth]{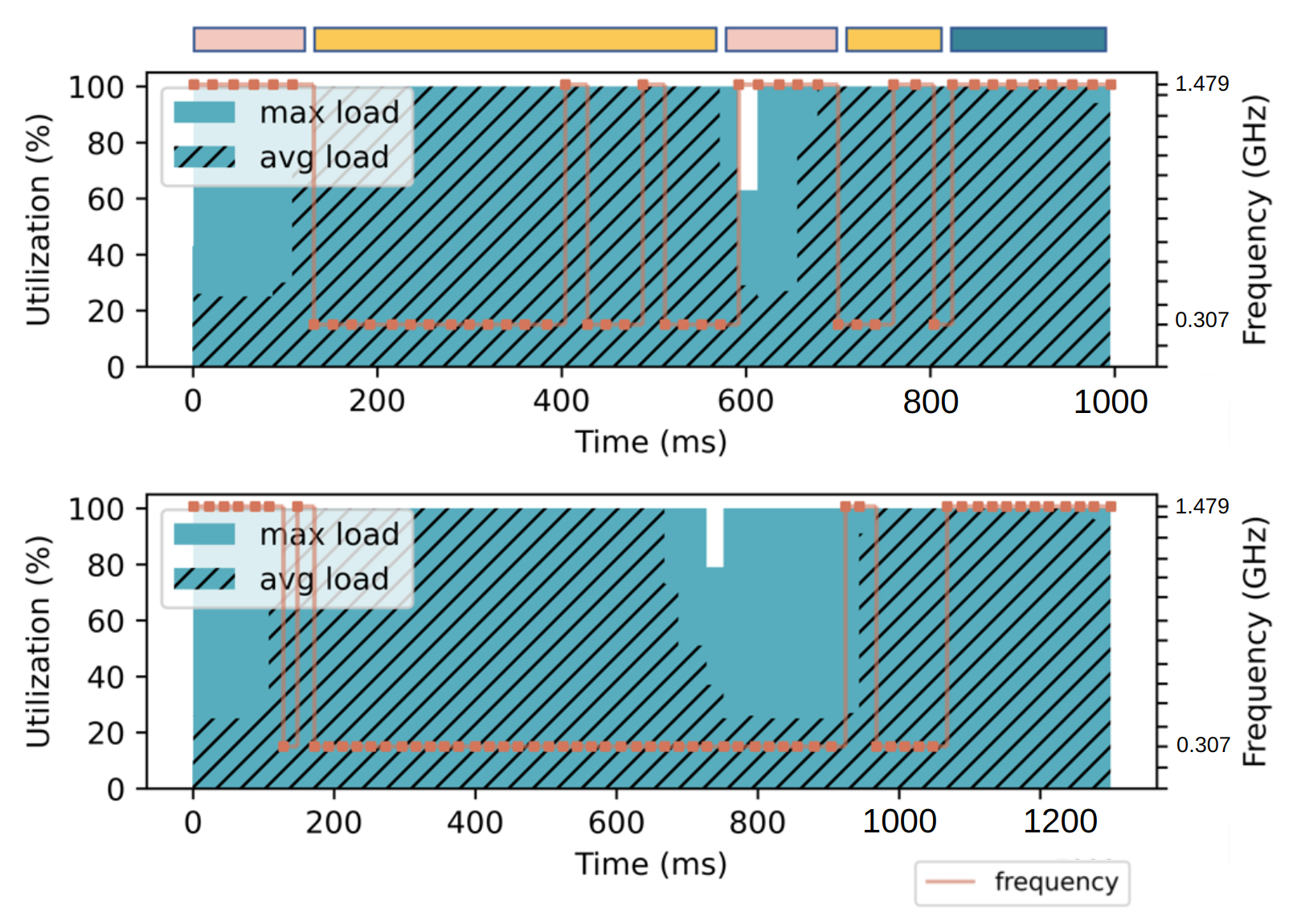}
	\caption{Frequency policy developed by proposed method on \textit{AudioRecog} with 1.0 s (top) and 1.3 s (bottom) deadline, and 0.6 s (top) and 0.9 s (bottom) microphone recording,  respectively.}
	\label{fig:e8}
\end{figure}

The visualization showing the frequencies, max utilization, and average utilization together at each observation point allows us to observe when the choices of low frequency or high frequency occur.
For Fig.~\ref{fig:e4} and Fig.~\ref{fig:e8}, we use the yellow bar to indicate the range with mostly low frequency and the pink bar to indicate the range with high frequency.
For the 0.6 s deadline, we observe that the model preferred to choose the low frequency in periods with high average utilization (yellow bars). In this case, the strategy chooses the high frequency for periods with low average utilization (pink bars).
For the 1.2 s deadline, a large number of low frequencies were adopted due to the reduced performance requirements.
For this workload, our proposed method showed the ability to use coarse-grained frequency support to fill the slack as well as to optimize overall CPU utilization.

The \textit{AudioRecog} workload is designed to test if our approach is able to  learn additional hidden abstract information: the existence of an internal slack.
At the highest speed (1.479 GHz), with 0.6 s microphone recording, it takes around 0.88 s to finish one request.
We assigned a 1.0 s deadline, which means there's about 0.12 s slack after the request is finished at the highest speed.
Also, there are around 0.28 s internal slack before the microphone recording is finished.
In the developed policy (Fig.~\ref{fig:e8} top), the proposed method exhibits the ability to preferentially select high average utilization periods to reduce the frequency (yellow bars).
At the same time, it drops the frequency heavily during the internal slack (yellow bar) and boosts it at the end (green bar), although both intervals exhibit an average utilization close to 100\%.
For 1.3 s deadline with 0.9 s microphone recording, the proposed method adjusts its policy (Fig.~\ref{fig:e8} bottom) with the change of performance settings.

\subsection{Energy Saving}

\begin{figure}
	\centering
	\includegraphics[width=0.5\textwidth]{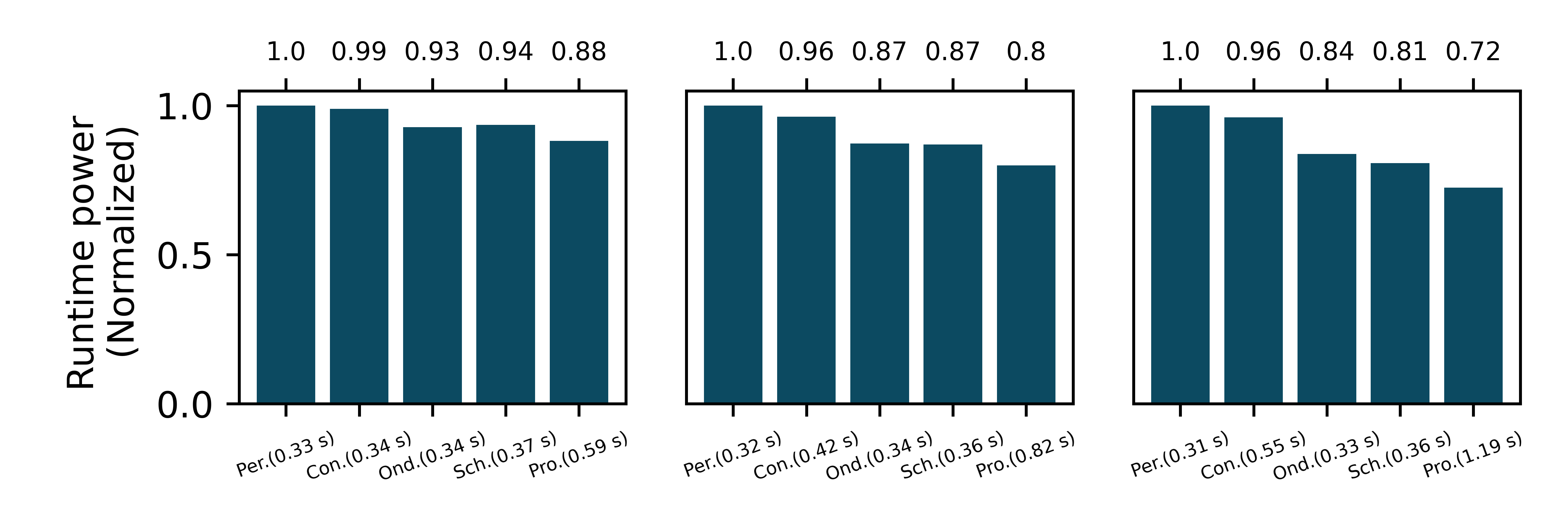}
	\caption{Power consumption on \textit{FaceRecog} with ddl = 0.6 s, 0.9 s, 1.2 s, from left to right respectively.}
	\label{fig:e5}
\end{figure}

\begin{figure}
	\centering
	\includegraphics[width=0.5\textwidth]{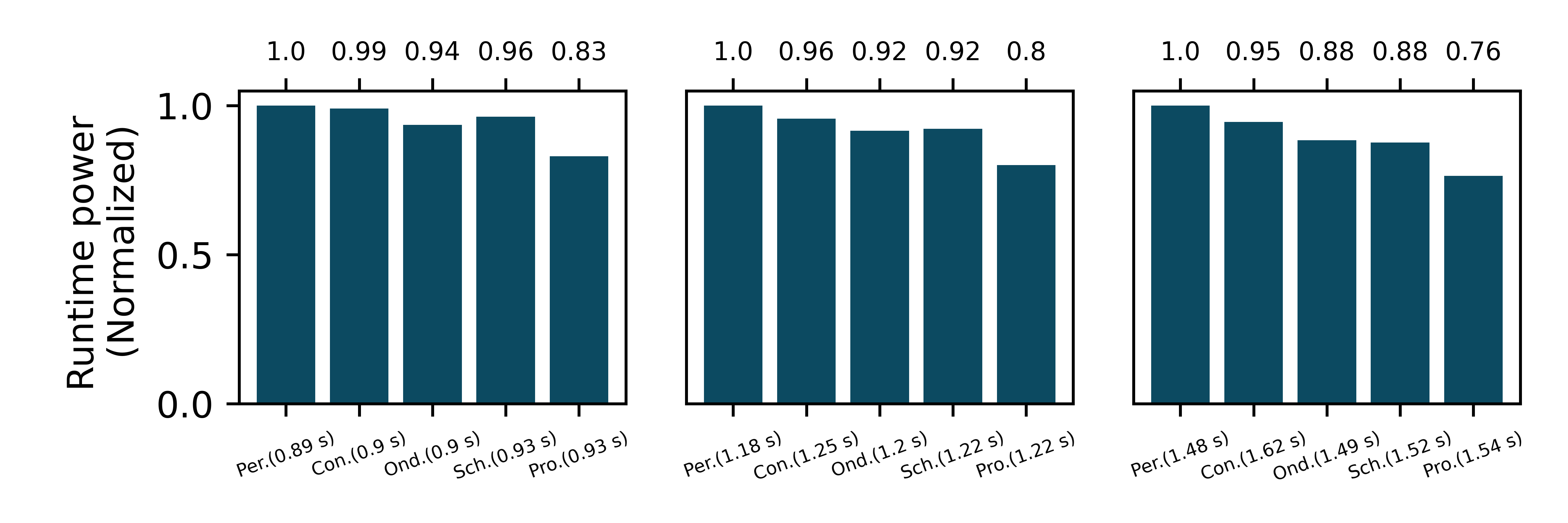}
	\caption{Power consumption on \textit{AudioRecog} with ddl = 1.0 s, 1.3 s, 1.6 s, from left to right respectively.}
	\label{fig:e9}
\end{figure}

\begin{figure}[t]
	\centering
	\includegraphics[width=0.5\textwidth]{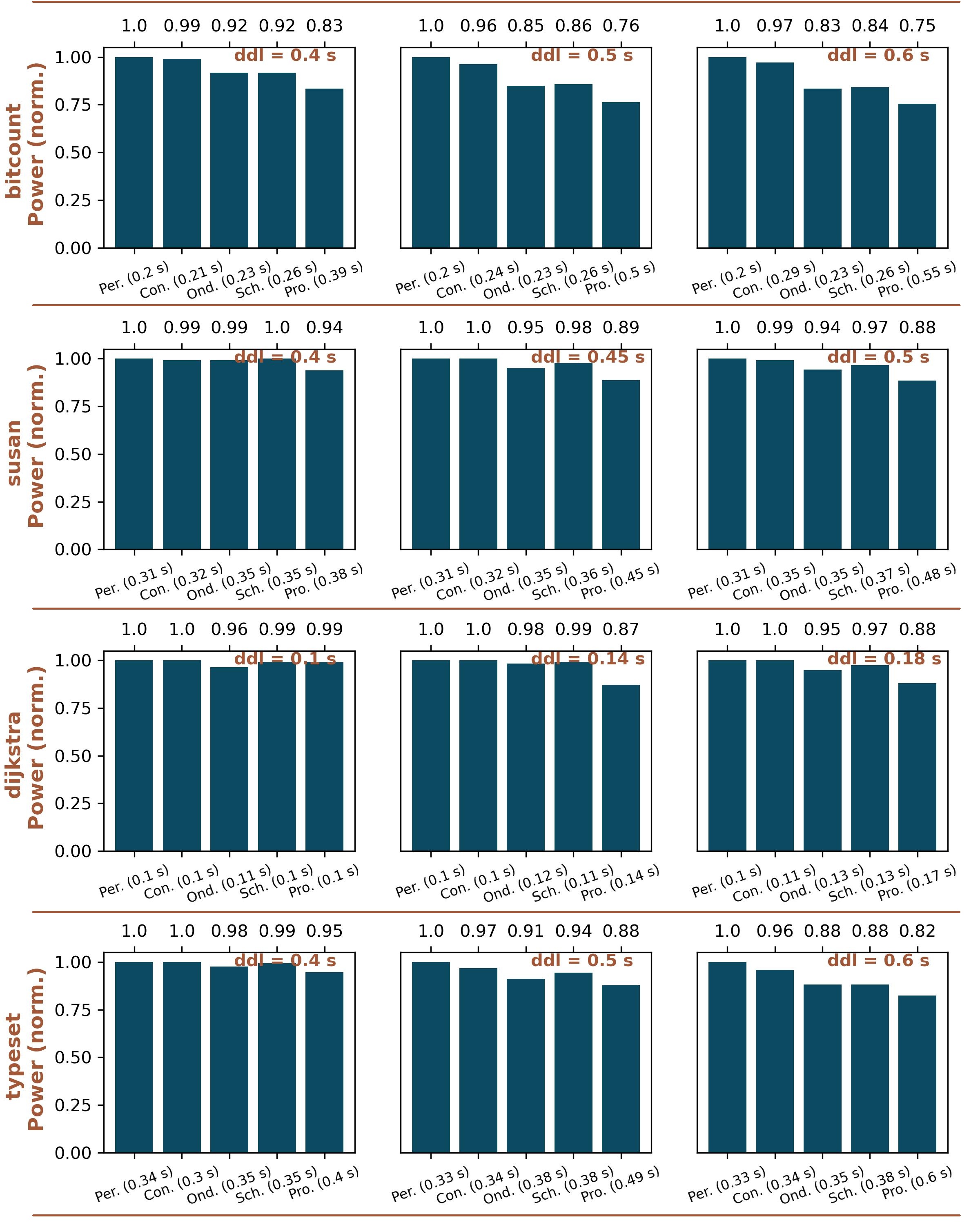}
	\caption{Power consumption on \textit{MiBench} workloads at various deadlines.}
	\label{fig:tmp}
\end{figure}

As shown in Fig.~\ref{fig:e5}, \ref{fig:e9} and \ref{fig:tmp}, 
our approach can self-develop DVFS policies to accommodate all chosen workloads (including self-designed workloads and workloads from MiBench) with different deadlines. The longer the deadline, the more energy-efficient our approach is compared to the built-in Linux approach.

In these figures, each row represents the results for one particular workload. Each column corresponds to one specific deadline. Each bar has the format of \textit{governor (running time)} to represent the average time to complete the workload under the given governor for the particular workload (indicated by row) and the particular deadline (indicated by column). 
To save chart space, we use the first three initials of the governor's name instead of its full name, e.g., \textit{Pro.} means \textit{our Proposed method}. The length of the bar indicates the normalized energy consumption. Each case's normalized energy consumption is shown at the top of the bars. We can observe that the proposed method consumes the least energy in almost all cases. Take the \textit{FaceRecog} example, the energy consumption is \textcolor{black}{88\%, 80\%, and 72\%} compared to the Performance governor for deadline = 0.6s, 0.9 s, and 1.2 s, respectively. For energy saving for other workloads and deadlines, please refer to Fig.~\ref{fig:e9} and \ref{fig:tmp}.

On \textit{Mibench} workloads, with performance slack (deadline - execution time under maximum execution speed) ranging from 0.04 s to 0.4 s, the proposed method can save 3\% - 11\% more energy compared to \textit{Ondemand}.
On \textit{FaceRecog}, with performance slack ranging from 0.27 s to 0.87 s, the proposed method can save 5\% - 14\% more energy compared to \textit{Ondemand}.
On \textit{AudioRecog}, with performance slack ranging from 0.11 s to 0.12 s, the proposed method can save 12\% - 14\% more energy compared to \textit{Ondemand}.
Note that for \textit{AudioRecog}, the energy-saving opportunity mostly comes from the internal slack.

\subsection{Inference Time Overhead}

The time overhead of the proposed method consists of constructing the system state and then reasoning about the state using the RL model.
We measure the time overhead by timing the execution of this block of code in the kernel.
The associated time overhead is influenced by two factors: The frequency at which the CPU is running when performing inference and the number of tasks that the relevant cores are processing concurrently.

For a fair measure, on Nvidia Jetson Nano 2GB Board, we run a task with 100\% CPU utilization on the core on which the DVFS governor is running and then measure the time overhead of running the proposed method at 1.479 GHz and 0.307 GHz, respectively.
For 1.479 GHz, we collected 1033 sets of data, and the average time overhead is 25.62 us.
For 0.307 GHz, we collected 1583 sets of data, and the average time overhead is 41.05 us.
We believe this is low enough overhead.

\textcolor{black}{\subsection{Deadline missing}}

\begin{figure}
	\centering
	\includegraphics[width=0.5\textwidth]{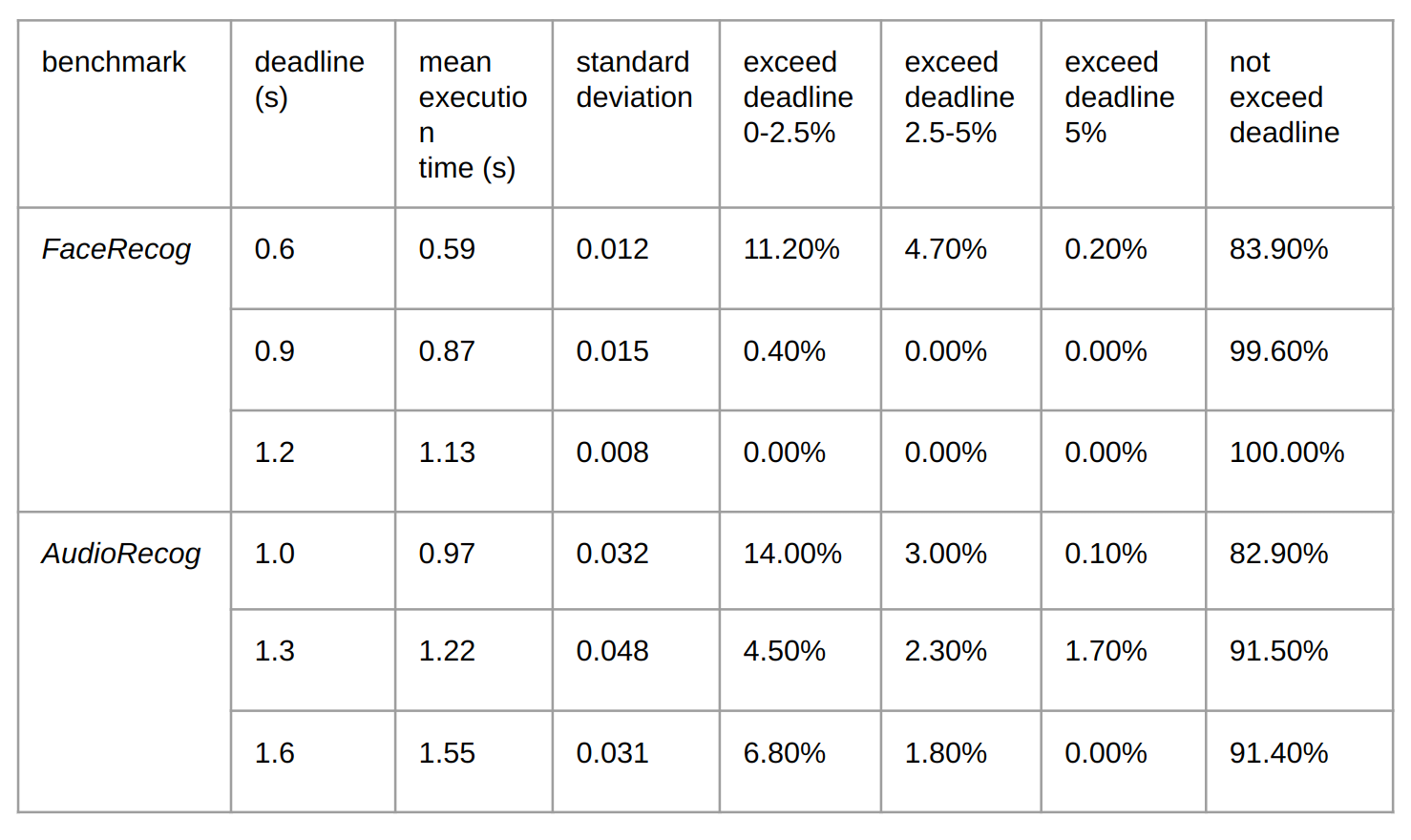}
	\caption{\textcolor{black}{Deadline missing test.}}
	\label{fig:deadline1000}
\end{figure}

\textcolor{black}{
For the two self-constructed workloads with three separate deadlines, we execute them 1000 times using a policy generated by reinforcement learning, and count the number of timeouts. The result is shown in figure \ref{fig:deadline1000}.
First, the time taken to execute workload per policy is very stable, as reflected by the low standard deviation (0.08-0.048).
When the average time taken to execute a workload is very close to a given deadline, it will miss the deadline in about 20\% of the cases.
For example, for \textit{FaceRecog}, at 0.6 deadline, the generated policy took 0.59 seconds to execute the workload once on average, and 16.1\% of the runs timed out.
We also note that the deadline timeout is only concentrated within 5\%, which is 0.03 s for a 0.6 s deadline.
Such a small number of timeouts can be considered as fluctuations caused by system events, and we believe that such results are good enough for soft real-time requirements.
}

\section{Compare with using RNN for end-to-end learning}
\label{compareRNN}
Our previous work \cite{zhou2021deadline} used Recurrent Neural Network (RNN) to provide end-to-end learning. This section discusses the difference between this work and the RNN-based model and explains why we want to use this encoding-based system.
For end-to-end learning, an RNN is used to process the observed workload sequences and generate encoded inputs into the Q Net. No human knowledge is involved in this approach.
In our previous work, for some simple workloads, on a hardware environment that supports ten fine-grained frequency support, the RNN approach can find a policy that sets the frequency as low as possible for the overall execution of the task without timeouts.

In this work, our target platform is embedded devices oriented to workloads containing three challenging features (coarse-grained f/V, unbalanced load distribution among cores, and workload having internal slack). Our testbed is the Jetson Nano Board.

We conduct a set of experiments here to compare the differences between the two schemes in this embedded device.  Our encoding method encodes a feature of length 6 from the workload sequence. In the scheme using RNN, we use a GRU (Gated Recurrent Unit) with input size 3 and hidden vector size 6, to automatically encode the workload sequence into a feature of length 6. Note that the feature lengths extracted by these two methods were set to be the same. The extracted features and candidate frequency action are processed, then input to a fully connected neural network (Q Net) of size 7-8-8-1. An evaluation of that frequency action results from the Q Net processing.
The training parameters and methods are the same for both schemes, as shown in table \ref{table:setting}.
The workload we use is \textit{FaceRecog} with a 0.6 s deadline.

\begin{figure}
	\centering
	\includegraphics[width=0.5\textwidth]{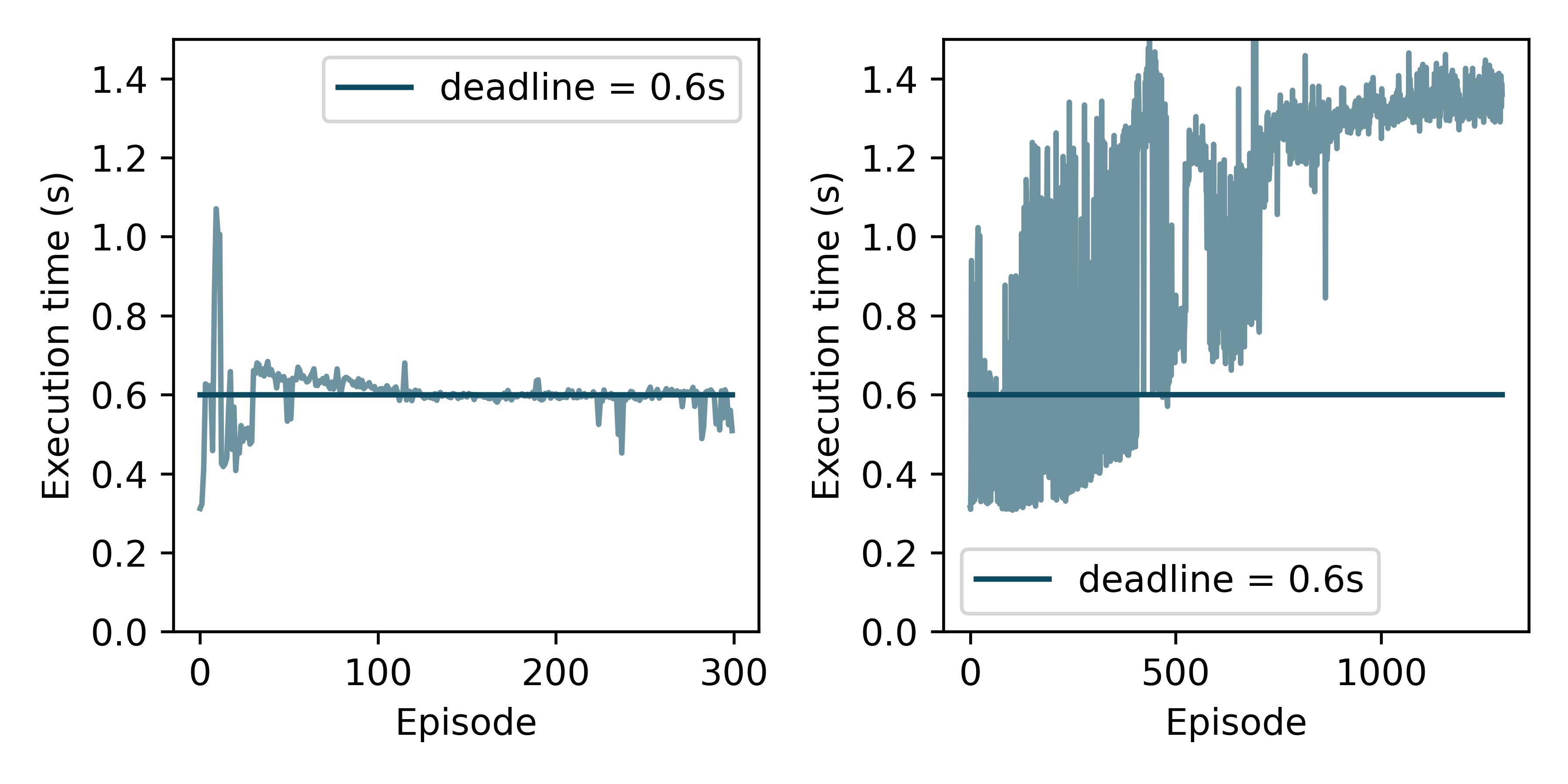}
	\caption{Training encoding-based learning (left, 300 episodes) and RNN-based learning (right, 1200 episodes) with large training pool size.}
	\label{fig:successed_encodingVSfailed_encoding}
\end{figure}

We first train the RNN-based learning the same way as the encoding-based method. For each new episode of experience collected, 64 sets of experiences are selected from each level of the experience pool and divided by reward level for training.
Fig. \ref{fig:successed_encodingVSfailed_encoding} shows the change in execution time of the workload during learning.
The encoding-based scheme demonstrates the ability to quickly sense the deadline (good strategies were developed with only 100 episodes) and maintain an understanding of requirements in subsequent training.
In contrast, despite learning 1200 episodes, the RNN-based approach shows no signs of approaching the deadline.

We further tune the training of the RNN scheme by changing the amount of the data used for training.
This time, 10 sets of experiences are selected from each experience pool bucket with level $>=$ 5, and 3 sets of experiences are selected from each bucket with level $<$ 5.

\begin{figure}
	\centering
	\includegraphics[width=0.5\textwidth]{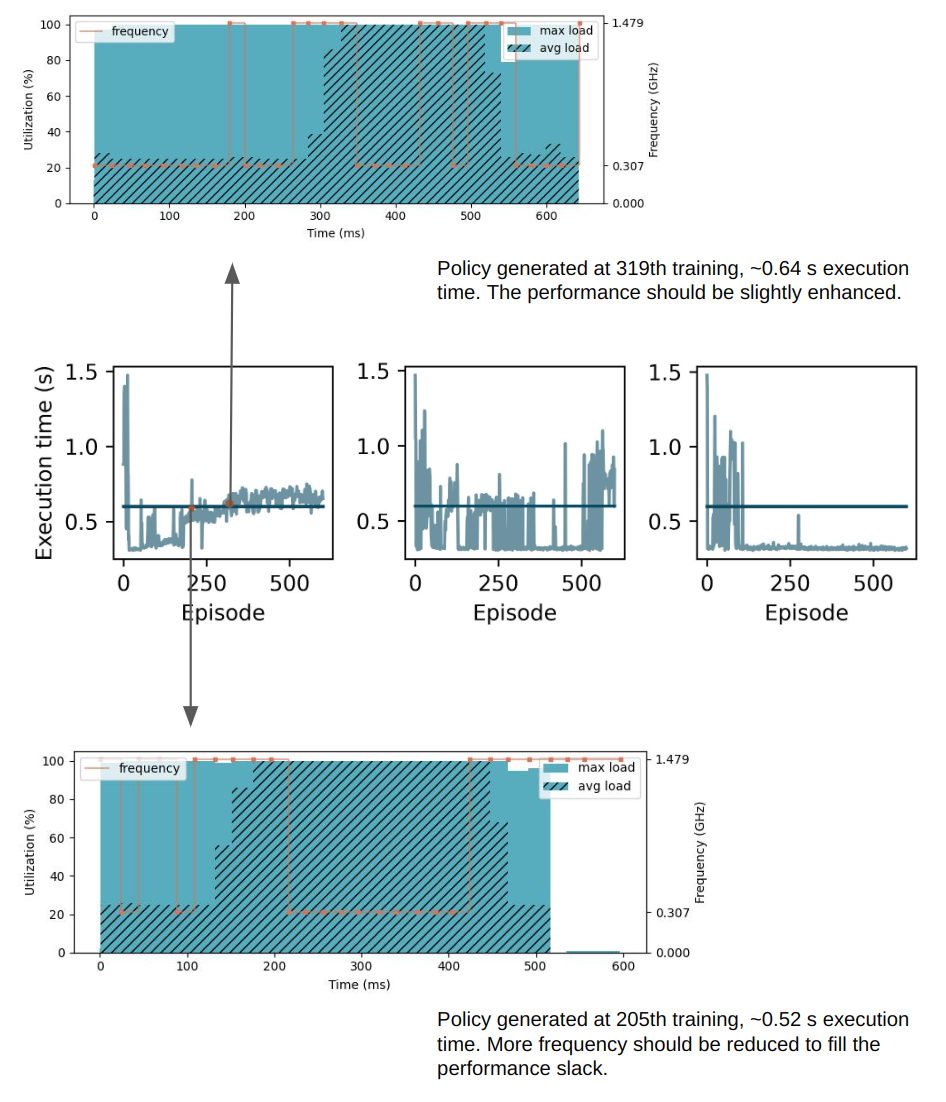}
	\caption{Training RNN-based method on \textit{FaceRecog} with 0.6 s deadline after tuning.}
	\label{fig:rnnTraining}
\end{figure}

\begin{figure}[t]
	\centering
	\includegraphics[width=0.5\textwidth]{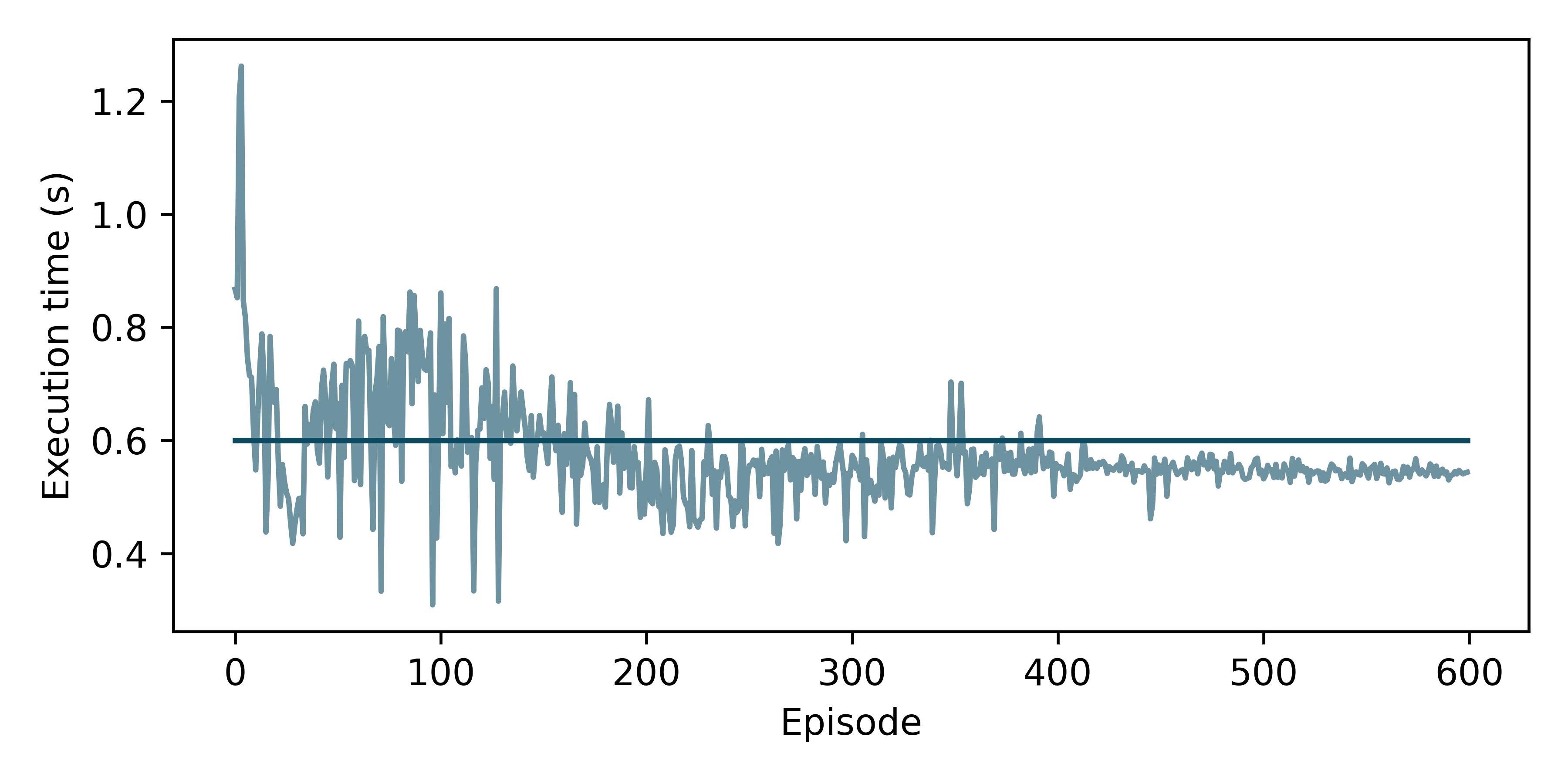}
	\caption{Training encoding-based method with smaller training pool size.}
	\label{fig:encoding_smaller_size}
\end{figure}

This time, for training the RNN-based method, we noticed much better results.
Fig. \ref{fig:rnnTraining} shows three learning curves.
One observation is that the RNN-based training method is \textbf{unstable}. Sometimes it shows a good perception of the deadline requirements (the figure on the left in Fig. \ref{fig:rnnTraining}), and sometimes a poorer perception (the figure on the right in Fig. \ref{fig:rnnTraining}).
In contrast, the learning curves generated by the proposed encoding-based method are consistently similar to Fig. \ref{fig:encoding_smaller_size}.

\begin{figure}
	\centering
	\includegraphics[width=0.5\textwidth]{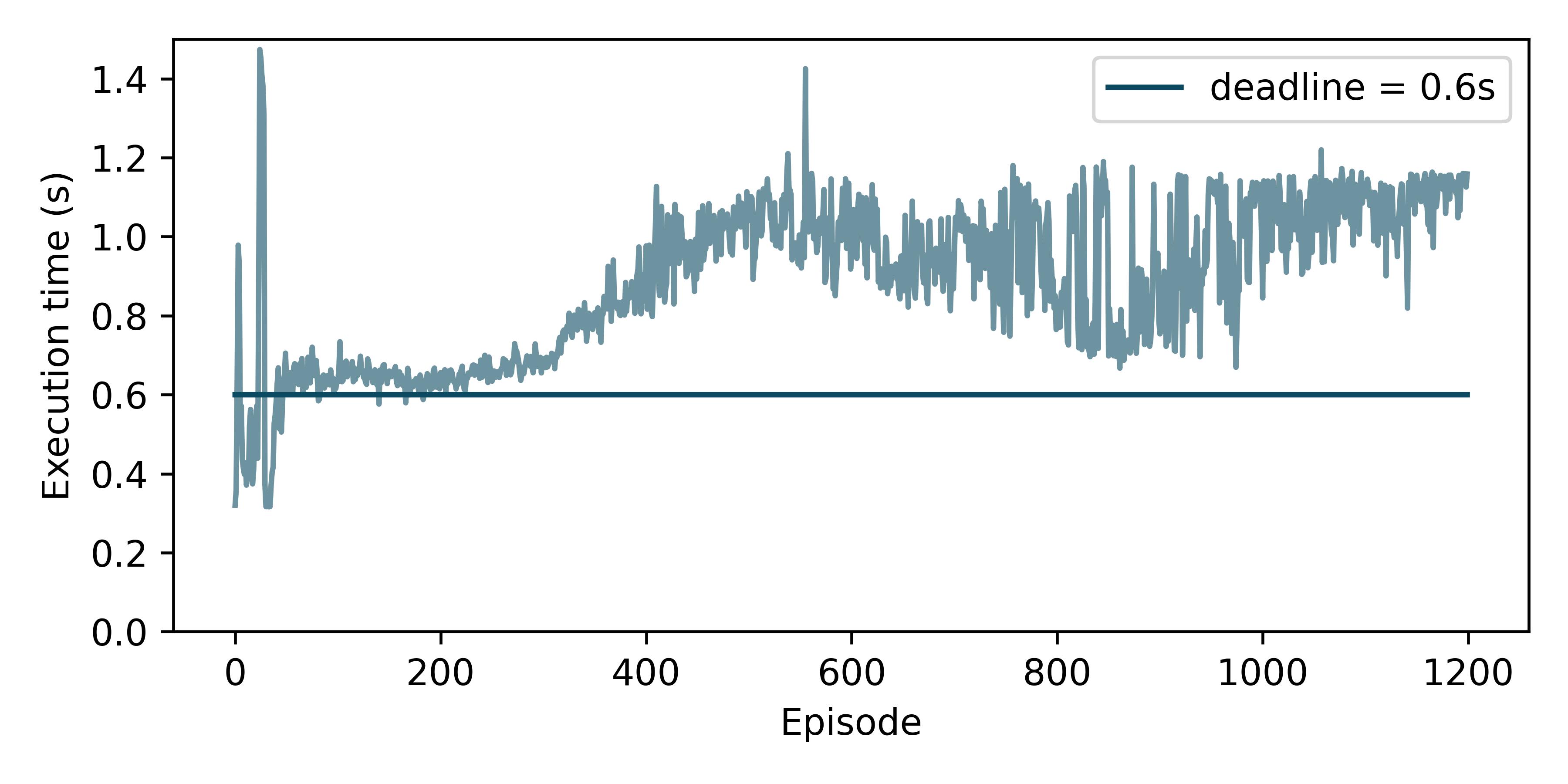}
	\caption{Training RNN-based method with 1200 episodes.}
	\label{fig:rnn_1200}
\end{figure}

\begin{figure}[t]
	\centering
	\includegraphics[width=0.5\textwidth]{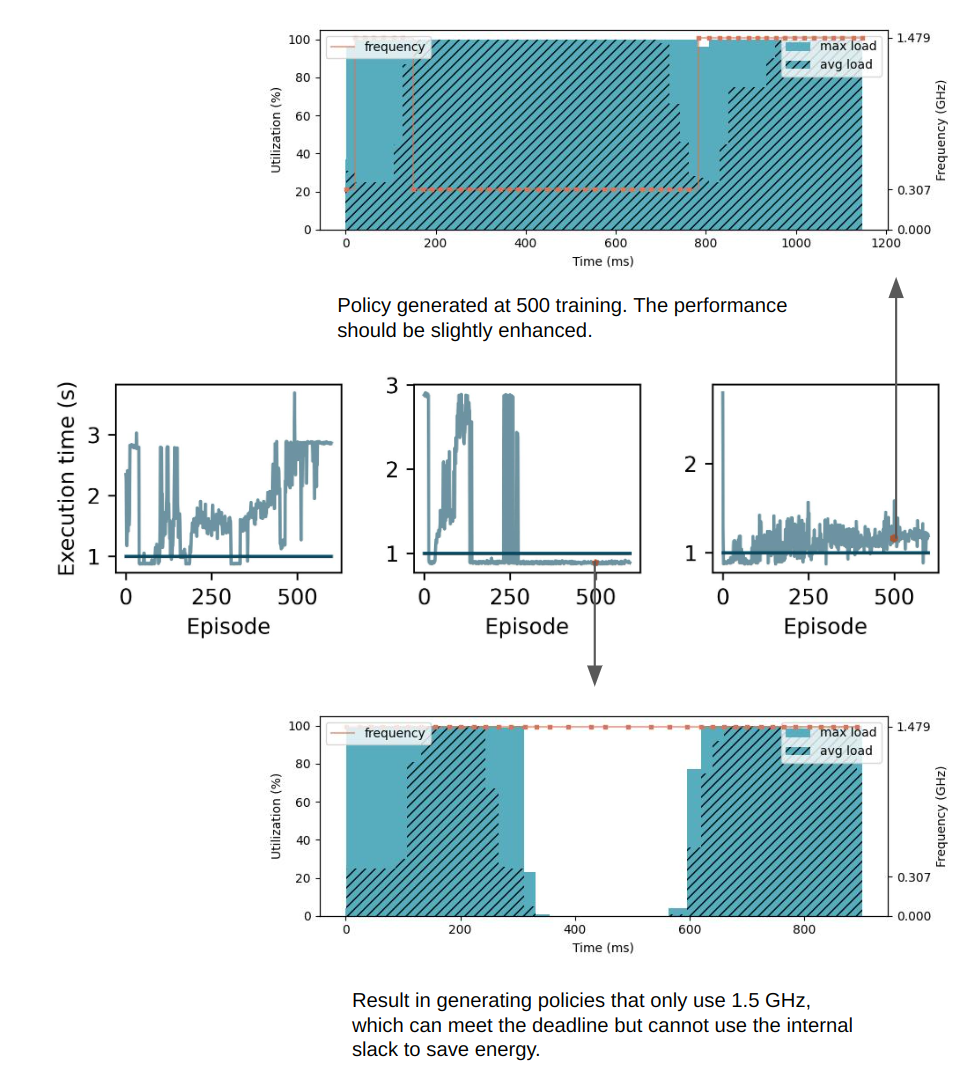}
	\caption{Training RNN-based learning on \textit{AudioRecog} with 0.6 s internal slack and 1.0 s deadline.}
	\label{fig:audio_recog_rnn_600}
\end{figure}

The RNN-based scheme can sometimes extract some strategies close to the ideal answer (the two policies on the left are visualized in Fig. \ref{fig:rnnTraining}). However, these models eventually could not meet the deadline requirement and kept exporting timeout policies. Extended training time would not help either (as shown in Fig. \ref{fig:rnn_1200}). 
We also trained the RNN-based method on \textit{AudioRecog}, and it shows similar patterns (Fig. \ref{fig:audio_recog_rnn_600}). 
For the challenging scenarios discussed in this paper, the RNN-based method demonstrates the ability to summarize knowledge to some extent, but not optimized. An example is shown in the middle figure in Fig. \ref{fig:audio_recog_rnn_600}, where the model continuously generates policies that always set the highest frequency (1.479 GHz) so as not to trigger a timeout but does not take advantage of the energy saving opportunity provided by the internal slack.

\begin{figure}
	\centering
	\includegraphics[width=0.5\textwidth]{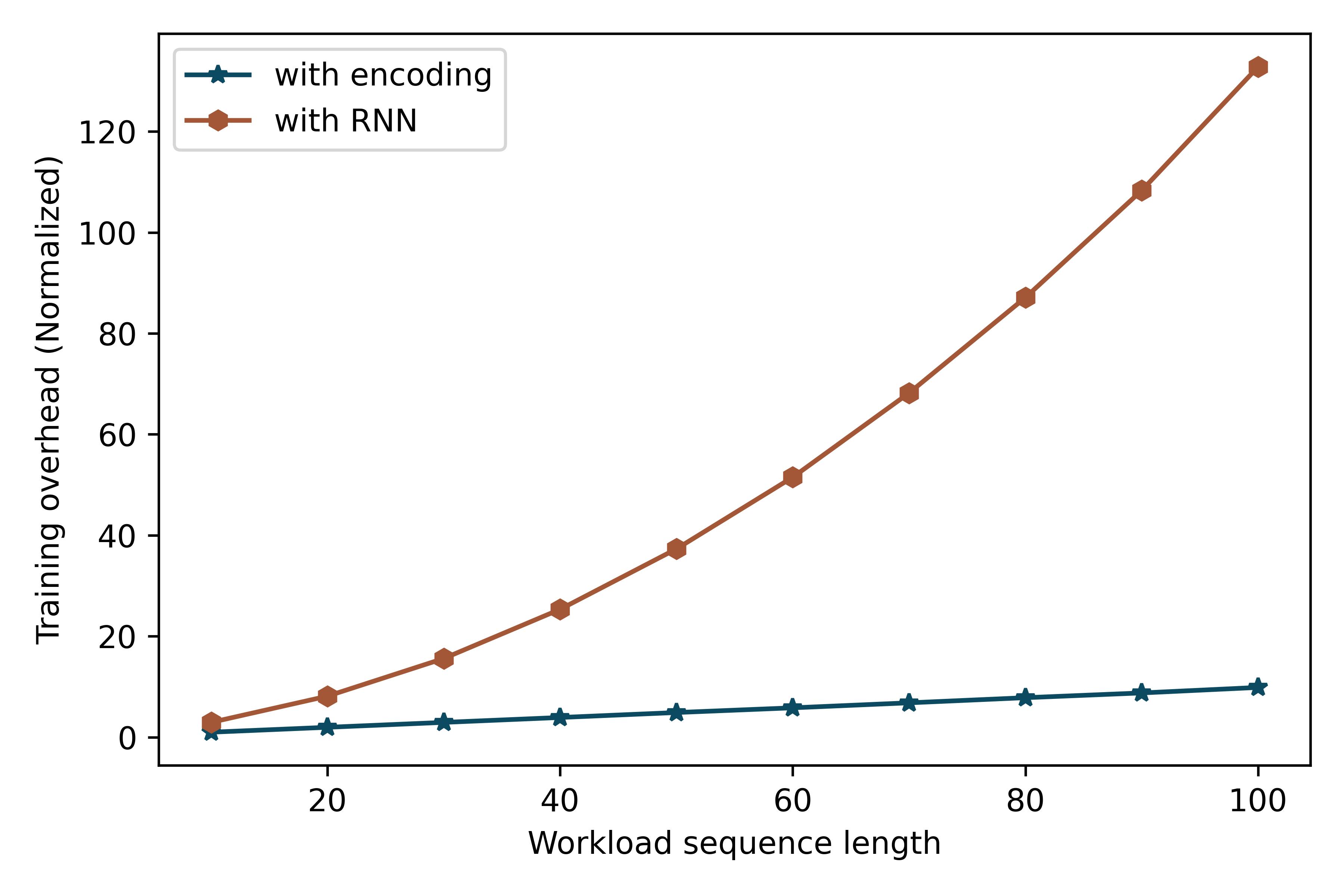}
	\caption{Training overhead increases as the length of workload increases.}
	\label{fig:overhead_rnn_vs_encoding}
\end{figure}

A problem that should not be overlooked is that during training, the RNN-based approach needs to process the complete observed sequence to obtain a feature. In contrast, the encoding-based approach can directly read the saved encoded features. This is because for RNN, each training changes its parameters and thus the extracted features, so the training requires reprocessing the sequence to get the new features. In contrast, the features generated by encoding schemes based on human knowledge are fixed and do not need to be recomputed.
For a workload sequence of length N, one complete learning would require RNN processing 1 + 2 + 3 + ... ...  + N = $\frac{N \times (N+1)}{2}$ times data. This overhead grows as the length of the workload grows. For the rule-based encoding scheme, this computational overhead does not exist.

For the network structure used in this experiment, the comparison of training time  is shown in Fig. \ref{fig:overhead_rnn_vs_encoding}.
When the workload length is 10, the training time of the RNN-based scheme is about 2.9 times higher than that is based on encoding.
When the workload length changes from 10 to 100, the RNN-based training overhead increases by a factor of about 45 and the encoding-based overhead only increases by a factor of about 9.8.
The training time with the expert-knowledge-based encoding scheme is much smaller and grows much more slowly.

\textbf{
The first reason we want to explore the encoding-based method over RNN-based is the low interpretability of neural network models}.
Up to now, the interpretability of neural networks is still poor. There is no obvious better way to improve a model except by adjusting the structure and making the experimental comparison. In addition, it is difficult to analyze the meaning of the data contained in the features encoded by the neural network from the human perspective. Therefore it is difficult to intervene in its learning process. \textbf{It is possible to achieve a better result by adjusting the RNN model structure and training method.} However, even if an ideal RNN structure is tuned for a workload, we cannot be sure that it can handle more complex features. We successfully processed simple workloads with RNN in a hardware environment that supports fine-grained frequency options in the above experiments. The same approach does not work well when dealing with the more challenging patterns discussed in this work. When the workload becomes even more complex, such as multi-tasking with multiple deadlines, the demands on the learning model will be higher. That is unless an extremely powerful model is fixed (which also implies a significant inference overhead), the user may need to adapt the model structure based on the workload, which is against the original purpose of wanting it to be adaptive.

Our encoding scheme can be seen as extracting valuable information from the raw data based on expert knowledge and then handing it over to a machine-learning model to map to the final control. In this approach, the main information extraction task is performed by human experts, thus reducing the reliance on machine learning structures. At the same time, such an approach provides a degree of interpretability: we can control and analyze the information perceived by the model by adjusting the encoded information. Our experiments demonstrate that the reinforcement learning model can generate good frequency control policies quickly and consistently based on the encoding scheme we designed.

\textbf{Another reason is the training overhead.}
As we discussed, the overhead of training an RNN-based model is much larger than that of training an encoding-based model. Our aim is to provide fast and lightweight learning on small devices. Excessive computational complexity prolongs learning time and is also a challenge for the device's heat dissipation capability. 

\section{Related Work}

Jung et al. \cite{jung2010supervised} used a Bayesian classifier, which is trained offline, to predict the performance and power dissipation of the processor for incoming tasks.
The features considered include task priority, queue occupancy, and arrival rate of the task.
A policy table calculated offline by dynamic programming was used to map the predicted state to V/f action.
Conradihoffmann et al. \cite{zhou2021deadline} used the Performance Monitoring Unit (PMU)s provided by the Cortex-A53 processor to offline analyze the correlation between performance counters (Bus Access for Memory Write, Read Alloc Mode, CPU Cycle, etc.) and target task's execution time.
An ANN model, which can learn online, was used to take in the selected features and predict task utilization.
A set of heuristic-based rules were designed to use the ANN prediction results to adjust the frequency while balancing the load.
Park et al. \cite{park2021interpretable} focused on developing highly interpretable solutions. They analyzed the tradeoff between precision and interpretability of various ML models on a dataset of mobile gaming workloads.
Tree-based linear models were finally selected and implemented to improve CPU/GPU utilization while achieving the target Frames-per-Second (FPS).
Das et al. \cite{das2015workload} used a statistical method to detect the application change point, along with an RL-based run-time manager and a hierarchical approach for V/f and thermal management.

% Wang et al. \cite{wang2016model} 
Ul et al. \cite{ul2015hybrid} used Q learning, a popular RL algorithm, to switch existing DVFS methods dynamically.
Based on the previous work, Ramegowda et al. \cite{ramegowda2021can} implemented and validated the hybrid DVFS method in various embedded devices running the Linux system.
Wang et al. \cite{wang2021online} used Double Q learning to explore the energy-performance optimization for both CPU core and uncore parts.
Specifically, they used the instruction per cycle (IPC), and the misses per operation (MPO) \cite{freeh2005using} as the state measurement of the environment and used $\frac{IPC^3}{W}$ as the reward to describe the tradeoff between energy and performance.
Although it was for high-performance computing, the goal was close to an embedded computing scenario: to be as energy-efficient as possible while meeting a global deadline.
Shafik et al. \cite{shafik2015learning} proposed a learning transfer-based adaption method, so the Q learning model, which only uses the CPU cycles in the last period as the system state, won't have to learn from scratch again when workload changes, thus reducing the convergence time.

\textcolor{black}{
In section \ref{compareRNN}, we compared the proposed temporal encoder-based approach with the RNN-based approach previously proposed \cite{zhou2021deadline}.
However, we have not been able to find other similar studies that can be fairly compared the results.
We next explain why based on three recent studies
\cite{conradihoffmann2021online} \cite{wang2021online} \cite{gupta2021dynamic}.
} 
\textcolor{black}{
The works done by Gupta et al. \cite{gupta2021dynamic} and Hoffmann et al. \cite{conradihoffmann2021online} can be seen as extending the architecture of Linux built-in methods by using more counters (Linux built-in methods use only utilization) to predict more events (Linux built-in methods assume the future period's utilization is the same as the past one), and designing corresponding rules to map the data to the frequency selection.
The biggest challenge of comparing to  Gupta's approach is that they did the experiments in a architecture simulator, and there is often a huge gap between the real system kernel environment and a simulator. \cite{conradihoffmann2021online} has done experiments on real systems and hardware. Unfortunately, there is no publicly available code and tuned parameters. And thus, it is difficult to reproduce the corresponding engineering implementation based on the paper description alone to have a fair comparison.
Wang et al. \cite{wang2021online} implemented their method in userspace and used user-state tools to collect information for states and rewards. Due to the overhead of user-level data collection, the work sets their frequency sampling rate at the sec-level, which would not work in our cases as the execution time of the workload we consider on embedded devices  is short. None of the above works consider temporal encoding to optimize frequency scaling by learning the task execution sequences.
}

\section{Conclusion}

% \textcolor{cyan}{
This work focus on energy saving for periodic systems constrained by the deadline on small devices. We identify three system patterns that may make Linux's built-in and similarly structured DVFS algorithms less effective.
We presented a reinforcement learning-driven DVFS governor using explicit temporal coding as input and experimented with it on an Nvidia Jetson Nano Board.  Our solution does not require an a priori model of the workload and devices architecture, which makes it practical and simple to deploy. This is similar to the long-established and well-tested Linux built-in systems. Our reinforcement learning method can derive a governor without introducing additional performance counters but can distinguish states with the same instance utilization through rapid profilings and learning with temporal encoding. The governor derived better addresses the three system patterns we identified and quickly adapts to six different applications and various performance requirements settings.

 Compared to the built-in Linux approach, the derived governor is able to leverage performance slack, save more energy, and place only a very small inference overhead burden.

\section*{}
This work is supported by the Natural Sciences and Engineering Research Council of Canada (NSERC).